\documentclass[journal, 10pt]{IEEEtran}
\usepackage{cite}
\usepackage{url}
\usepackage{hyperref}
\hypersetup{colorlinks=true, linkcolor= red, citecolor=red, urlcolor=blue}
\usepackage{bm,amssymb}
\usepackage{amsmath}
\usepackage{array}
\usepackage{amsthm}
\usepackage{mathptmx}
\usepackage[linesnumbered,ruled,vlined]{algorithm2e} 
\usepackage{float}
\usepackage{algpseudocode}
\usepackage{xcolor}
\usepackage{graphicx}
\usepackage{mathtools, cuted}
\allowdisplaybreaks

\newcolumntype{K}[1]{>{\centering\arraybackslash}p{#1}}
\DeclareMathOperator{\Tr}{Tr}
\usepackage{setspace}

\makeatletter
\newcommand{\HinTwoColAlg}{\let\@latex@error\@gobble}
\makeatother
\begin{document}
\title{Mobility, Communication and Computation Aware Federated Learning for Internet of Vehicles}

\author{\IEEEauthorblockN{Md Ferdous Pervej\IEEEauthorrefmark{1}\IEEEauthorrefmark{3}, Jianlin Guo\IEEEauthorrefmark{1}, Kyeong Jin Kim\IEEEauthorrefmark{1}, Kieran Parsons\IEEEauthorrefmark{1}, Philip Orlik\IEEEauthorrefmark{1}, Stefano Di Cairano\IEEEauthorrefmark{1}, Marcel Menner\IEEEauthorrefmark{1}, Karl Berntorp\IEEEauthorrefmark{1},
Yukimasa Nagai\IEEEauthorrefmark{2}, and Huaiyu Dai\IEEEauthorrefmark{3}}\\
\IEEEauthorblockA{\IEEEauthorrefmark{1}Mitsubishi Electric Research Laboratories (MERL), Cambridge, MA, 02139 USA\\
\IEEEauthorrefmark{1}Emails:{\tt\{guo, kkim, parsons, porlik, dicairano, menner, berntorp\}@merl.com}
\IEEEauthorrefmark{2}Information Technology R\&D Center, Mitsubishi Electric Corporation, Kamakura, Kanagawa 2478501 Japan\\
\IEEEauthorrefmark{3}Department of Electrical and Computer Engineering, NC State University, Raleigh, NC, 27606 USA\\ 
\IEEEauthorrefmark{3}Emails:{\tt\{mpervej, hdai\}@ncsu.edu} 
\vspace{-0.25in}
}
\thanks{This work was done while Md Ferdous Pervej was working at MERL as an intern.}
} 

\maketitle

\begin{abstract}
While privacy concerns entice connected and automated vehicles to incorporate on-board federated learning (FL) solutions, an integrated vehicle-to-everything communication with heterogeneous computation power aware learning platform is urgently necessary to make it a reality.
Motivated by this, we propose a novel mobility, communication and computation aware online FL platform that uses on-road vehicles as learning agents.
Thanks to the advanced features of modern vehicles, the on-board sensors can collect data as vehicles travel along their trajectories, while the on-board processors can train machine learning models using the collected data.
To take the high mobility of vehicles into account, we consider the delay as a learning parameter and restrict it to be less than a tolerable threshold. 
To satisfy this threshold, the central server accepts partially trained models, the distributed roadside units (a) perform downlink multicast beamforming to minimize global model distribution delay and (b) allocate optimal uplink radio resources to minimize local model offloading delay, and the vehicle agents conduct heterogeneous local model training.
Using real-world vehicle trace datasets, we validate our FL solutions.
Simulation shows that the proposed integrated FL platform is robust and outperforms baseline models. With reasonable local training episodes, it can effectively satisfy all constraints and deliver near ground truth multi-horizon velocity and vehicle-specific power predictions.
\end{abstract}


\IEEEpeerreviewmaketitle

\section{Introduction}
Modern vehicles are packed with various on-board sensors to sense diversified data for fulfilling higher automation levels.
On the one hand, data collected by individual vehicles may be non-IID (independent and identically distributed) and contain imperfections. 
Training independent a machine learning (ML) model solely on individual vehicle data may lead to non-robustness and prediction may not suffice in some strict applications.
In addition, these independently sensed data may not avail predicting a global task with data features/labels that these sensors have never observed. 
Moreover, some vehicles may not have enough computation resources to train a ML model. 
On the other hand, sending raw data to a central server is impractical due to extensive privacy threats and enormous communication overhead.
As such, the above-mentioned issues may very well become a bottleneck to delivering Society of Automotive Engineers (SAE) level $5$ automation on the road.

The recent advances of privacy-preserving federated learning (FL) can bring a solution to this bottleneck.
FL is an advanced ML technique that allows training ML models locally based on the trainer's local data \cite{mcmahan2017communication}. 
Therefore, it ensures user privacy protection and can effectively reduce communication overhead. Most importantly, FL incorporates data features in collaborative heterogeneous datasets, which allows robust traffic model training by eliminating data imperfection contained in individual dataset and ensures a faster convergence \cite{9311906}. 
Robust traffic models can be distributed to vehicles for their tasks at any location on the road.
Leveraging FL, modern vehicles can not only train traffic models on their on-board processors but can also keep their privacy unscathed.
Therefore, an integrated FL platform with advanced vehicle-to-everything (V2X) communications seems inevitable for the rising demand on higher automation.

While FL with vehicular agents can indeed bring manifold benefits, it also raises new challenges. Perhaps, two major concerns are ($1$) the high mobility and ($2$) the delay. Due to the high mobility, the time a vehicle connects to a connection point may be short.
As a result, the long delay may not be acceptable. 
Even though FL can comparatively reduce the communication overhead, model distribution (delivering the global model from the central server to vehicles) and offloading (uploading local models from vehicles to the central server) still take time.
Furthermore, local model training using the limited computation power of the on-board processors may be a dominant contributor to the delay.
As such, hyper-parameters such as total global training rounds (communication rounds between the central server and the vehicle agents), total local training iterations and vehicle agent selection are critical for swift model training and updating.

Advanced ML techniques are widely used to predict different traffic metrics such as traffic flow, congestion and trajectory. 
Nevertheless, independent learning may not be sufficient due to data heterogeneity, non-IID distribution and imperfection.
Therefore, crowdsource-based collaborative learning such as FL is luring researchers with its ensured privacy protection and use of diverse data samples of the learning agents.
There exist works  
that leverage FL in the vehicular environment. However, most of these studies have either not considered the mobility of the vehicles or not examined the underlying heterogeneous computation power of the vehicles.
In \cite{zeng2021}, the authors proposed a multi-task FL framework to predict average vehicle velocity using stationary agents.
The authors in \cite{kong2021fedvcp} presented a FL based vehicular cooperative positioning architecture to model the location of the vehicles.
However, this work does not consider model training delay and computation power heterogeneity. In addition, the authors assumed data augmentation over vehicle-to-vehicle (V2V) links that raises privacy concerns and communication overhead.
an agent-edge-cloud FL architecture was proposed in \cite{liu2020client} where the authors intended to minimize communication latency between the server and the agent. However, the agents are stationary.
A similar architecture was used in \cite{zhou2021two} for object detection in vehicular environment with a focus on context instead of mobility, communication and computation. 

In a vehicular environment, mobility is a nagging concern. 
This work proposes an integrated vehicular FL platform that takes account of the vehicle mobility and the distributed model training delay including local model training time, model queuing time, and model transmission time.

The contributions of this paper are summarized as follows:
\begin{itemize}
    \item We present a two-tier hierarchical vehicular FL platform to address the short connectivity issue caused by the high mobility to achieve robust traffic model training. 
    \item Instead of applying traditional FedAvg based FL, we propose an extended FedProx \cite{MLSYS2020_38af8613} based FL to
    incorporate vehicle mobility, communication delay, queuing delay, model training delay and data heterogeneity.
    \item In our FL platform, each roadside unit (RSU) solves a multicast beamforming problem to maximize the minimum sum rate of its associated agents for model distribution in the downlink. 
    To offload the locally trained models, each RSU solves a complex combinatorial problem to allocate optimal radio resources to the learning agents.
    \item While any ML task can be solved using the proposed online FL platform, we predict multi-horizon vehicle-specific power (VSP) requirements by exploiting real-world vehicle datasets. 
\end{itemize}

\begin{table}[!t]
\centering
\fontsize{8}{8}\selectfont
\caption{\textbf{Table of Notation}}
\begin{tabular}{|K{0.9cm}|K{6.9cm}|}
\hline
    \textbf{Symbol} & \textbf{Description} \\ \hline
    $v$ & A vehicle in the on-road vehicle set $\mathcal{V}$ \\ \hline
    $b$ & A roadside unit (RSU) in the distributed RSU set $\mathcal{B}$ \\ \hline
    $\mathcal{V}_k$ & Set of vehicles selected as learning agents in global training round $k$ \\ \hline
    $\mathcal{V}_k^{b}$ & Set of vehicle agents associated with $b$ in global training round $k$ \\ \hline
    $\mathbf{w}$ & Global machine learning model \\ \hline
    $\mathbf{w}_k$ & Global machine learning model in global training round $k$ \\ \hline
    $\mathbf{w}_k^{v}$ & Machine learning model trained by vehicle $v$ in global round $k$\\ \hline
    $\Omega$ & Total network bandwidth for vehicle-RSU communications \\ \hline
    $\Omega_b$ & Network bandwidth assigned to RSU b \\ \hline
    $z_b$ & A physical resource block (pRB) in RSU $b$'s pRB set $\mathcal{Z}_b$\\ \hline 
    $\mathbf{g}_{b}^z$ & Downlink multicast beamforming vector of RSU b over
pRB $z_b$ \\ \hline 
    $\Gamma_{b,z}^{v, \text{dn}}$ & Downlink SNR from RSU $b$ to vehicle $v$ over pRB $z_b$ \\ \hline 
    $\Gamma_{b,z}^{v, \text{up}}$ & Uplink SNR from vehicle $v$ to RSU $b$ over pRB $z_b$\\ \hline
    $d_{v}^{\text{down}}$ & Downlink global model transmission time from RSU $b$ to vehicle $v$ \\ \hline
    $d_{v}^{\text{up}}$ & Uplink local model transmission time from vehicle $v$ to RSU $b$ \\ \hline
    $d_v^{\text{cmp}}$ & The amount of time for vehicle $v$ to locally train  model $\mathbf{w}_k$ \\ \hline
    $d_v^{q, \text{up}}$ & Uplink queuing time of model $\mathbf{w}_k^v$ at vehicle $v$ \\ \hline
    $\mathrm{d}_v^{\text{tot}}$ & Total delay at vehicle $v$ in a global training round \\ \hline
    $\mathrm{d}^{\text{thr}}$ & Time threshold for all $v \in \mathcal{V}_k$ to finish model training \& offloading \\ \hline
    $L_v$ & Number of local training iterations at vehicle $v$ \\ \hline
    $\mathcal{D}_k^{v}$ & Dataset of vehicle $v$ in a global training round $k$ \\ \hline
    $P_b^z$ & RSU $b$'s transmission power over pRB $z_b$, $P_b^z \le P_b^{max}$ \\ \hline
    $P_{v}^b$ & Uplink transmission power from vehicle $v$ to RSU $b$\\ \hline
\end{tabular}
\label{notation}
\end{table}

\section{System Model and V2X Communication Model}
\label{Sec_System_Model}
This section presents our system model and communication model for the proposed vehicular FL platform.
Moreover, for convenience, we summarize the main notations in Table \ref{notation}.

\subsection{System Model}
We consider an Internet of Vehicles (IoV) network that consists of a central server, multiple RSUs and on-road vehicles. 
Denote the vehicle set and the RSU set by $\mathcal{V} =\{v\}_{v=1}^V$ and $\mathcal{B} = \{b\}_{b=1}^B$, respectively.
At a high level, the learning process progresses as follows. The central server first selects vehicular agents and distributes a global model, parameterized by its weight $\mathbf{w}$, to the selected agents. Upon receiving $\mathbf{w}$, an agent $v$ trains the received model using its local data and offloads the updated model weight $\mathbf{w}^{(v)}$ to the server. The server then aggregates all agents' local models to build an updated global model. The training process continues until the global model reaches an expected level of precision.

\begin{figure}
\centering
\includegraphics[width=0.45 \textwidth]{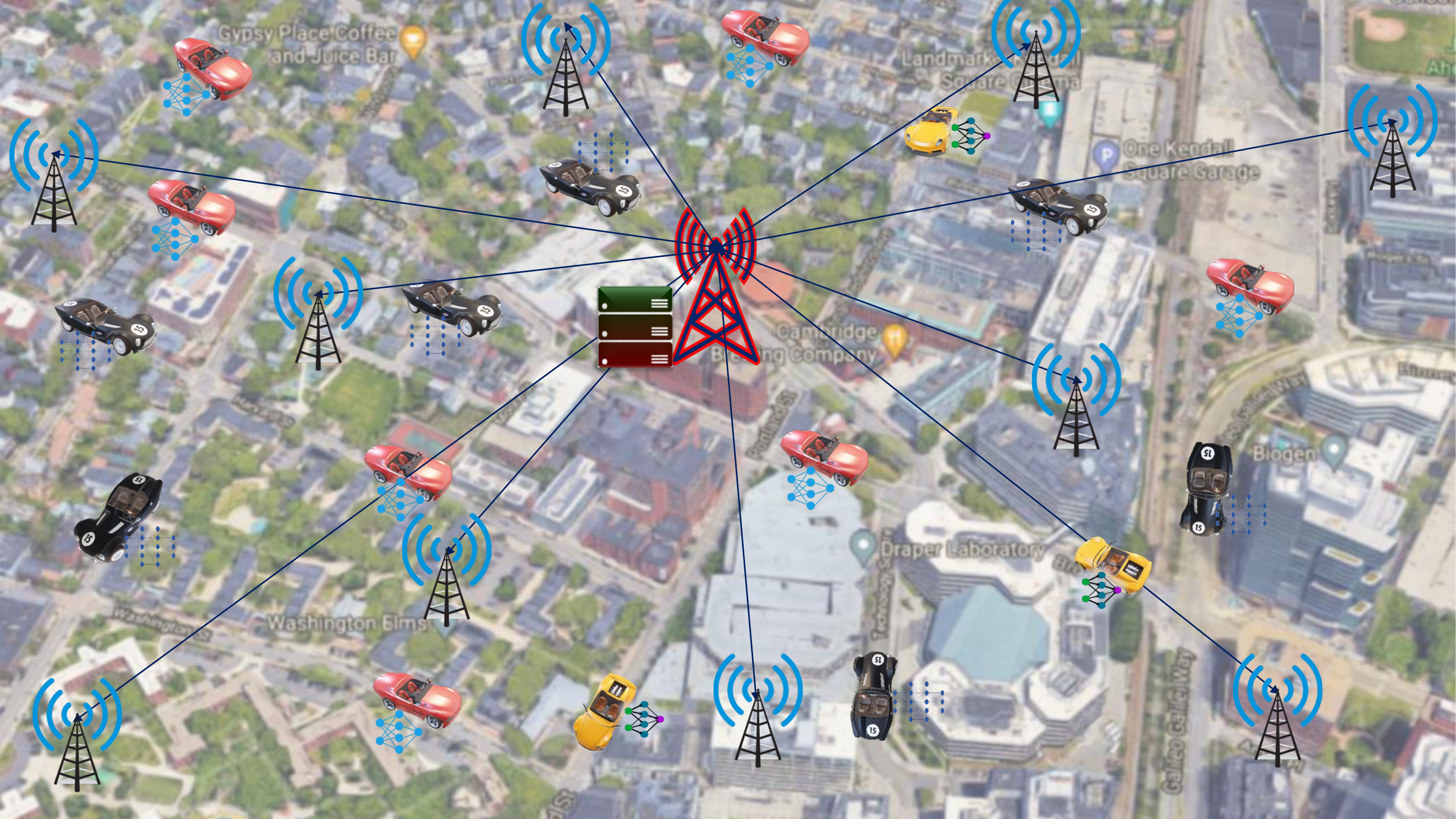} 
\caption{Proposed vehicular FL architecture for IoV}
\label{sys_mod_fig}
\end{figure}

To perform such distributed model training in a vehicular environment, an efficient infrastructure-to-vehicle (I$2$V) / vehicle-to-infrastructure (V$2$I) communication platform is needed for sharing $\mathbf{w}$ and $\mathbf{w}^{(v)}$s.
Therefore, this paper proposes a practical dense heterogeneous network architecture, where multiple RSUs are deployed over a considered region of interest (RoI) to connect vehicles as shown in Fig. \ref{sys_mod_fig}.
These RSUs are connected to a central server with high-speed wired links.
The vehicles travel on the roads and get connected to the network via these RSUs.

The motivation for adopting the proposed system model is three-folded. 
Firstly, the on-board computation power of vehicles is limited. It may take a longer time to finish the local model training. 
Therefore, vehicles may move away from the RSUs used by the server to distribute the global model.
The proposed system model enables vehicles to be reasonably covered by other RSUs while they perform local model training.
Secondly, it aims to ensure V$2$I connectivity in a larger area for comprehensive observations and robust model training.
Thirdly, it may not be possible to establish direct vehicle-server communication links due to the limited coverage of wireless links.
The proposed system model can not only ensure vehicle data privacy, but also allow the vehicles to accumulate more observations.
It can be used by various vehicular applications that highly rely on observations of vehicles, e.g., the adaptive advanced driver-assistance systems (ADAS).

\subsection{Vehicular Communication Model}
Note that due to high-speed wired links between the server and RSUs, the server-RSU communication delay is negligible and ignored in this work.
Assume each vehicle has a single antenna, while each RSU has $n_b$ antennas.
Besides, let the total network bandwidth, for the vehicle-RSU communications, be $\Omega$ Hz.
We divide the radio resource to the RSUs so that the consecutive RSUs have independent radio resources.
We reuse the radio resources by ensuring that the reusing RSUs are far away from each other. 
Furthermore, we divide the allocated radio resources into orthogonal physical resource blocks (pRBs).
Without loss of generality, let us denote the radio resource of RSU $b$ by $\Omega_b$ and denote the pRB set of $b$ by $\mathcal{Z}_b=\{z_b\}_{z_b=1}^{Z_b}$.
We consider that all pRBs have the same size $\omega$.
Moreover, with an appropriate frequency reuse pattern and orthogonal partitioning of the system bandwidth, there is no interference in our proposed system model.

The proposed network operates in time division duplex (TDD) mode and can exploit channel reciprocity. The channel is assumed to be quasi-static block fading, i.e., it remains static within a block but varies across different blocks.
Denote the wireless channel between the RSU $b$ and vehicle $v$ over pRB $z_b$ by $\mathbf{h}_{b,z}^{v} \in \mathbb{C}^{n_b \times 1}$.
Moreover, we model the channel as $\mathbf{h}_{b,z}^{v}=\sqrt{\zeta}_b^v \tau_b^v \breve{\mathbf{h}}_{b,z}^v$, where $\sqrt{\zeta}_b^v$, $\tau_b^v$, and $ \breve{\mathbf{h}}_{b,z}^v$ are large scale fading, log-Normal shadowing, and small scale fading channel response, respectively. We model the pathloss following $3$GPP's \textit{UMa} model \cite{3GPP_TR_38_901} as suggested in \cite{3GPP_TR_38_886}.
For simplicity, we assume RSUs and vehicles can mitigate Doppler spread perfectly.

In 3GPP based IoV network, the RSUs can be connected with each other via the $\mathrm{X_n}$ interface. 
In addition, we consider a mobility model where each agent calculates the reference signal received powers (RSRPs) from its serving RSU and adjacent RSUs.
Once the A$3$ event \cite{sesia2011lte} is triggered, the agent reports the measurements to the serving RSU.
The serving RSU then handovers the agent to the target RSU \cite{sesia2011lte}.
As such, an agent is always associated with only one RSU, i.e., $\sum_{b=1}^B a_v^b=1$, $\forall v \in \mathcal{V}$, where $a_v^b \in \{0,1\}$ is an indicator function that takes value $1$ when vehicle $v$ is associated to RSU $b$.

\subsubsection{Downlink Communication Model}
In the downlink, as a RSU transmits the same model to all of its associated agents, each RSU $b$ can multicast the ML model $\mathbf{w}$\footnote{In a practical wireless network, the RSU can use the multicast broadcast single frequency network (MBSFN) subframe \cite{sesia2011lte} to broadcast the model.}. We assume that each RSU has fixed pRBs to multicast the ML model.
Denote the downlink multicast beamforming vector of RSU $b$ over pRB $z_b$ by $\mathbf{g}_{b}^z \in \mathbb{C}^{n_b \times 1}$.
Then, the received signal at vehicle $v$ over pRB $z_b$ is expressed as
\begin{equation}
\label{downlink_signal_at_v}
    y_{b,z}^{v, \text{dn}} = \sqrt{P_{b}^{z}} a_v^b ({\mathbf{h}_{b,z}^v})^H \mathbf{g}_{b}^z s_b^v  + \eta,
\end{equation}
where $P_b^z$ is $b$'s transmission power over pRB $z_b$, $s_b^v$ is the unit power transmitted data symbol of $b$ intended for vehicle $v$ and $({\bm x})^H$ denotes the conjugate transpose. 
Moreover, $\eta \sim CN(0, \sigma^2)$ is the circularly symmetric complex Gaussian noise with zero mean and variance $\sigma^2$.

To this end, we calculate the downlink signal-to-noise ratio (SNR) over pRB $z_b$ as follows:
\begin{equation}
    \Gamma_{b,z}^{v, \text{dn}} ={P_b^z  | ({\mathbf{h}_{b,z}^v})^H \mathbf{g}_{b}^z  |^2} /{\omega \sigma^2}.
\end{equation}
When the global FL model is distributed from the server, we calculate the downlink rate at vehicle $v$ from RSU $b$, over all pRB $z_b \in \mathcal{Z}_b$, as
\begin{equation}
    C_v^{\text{dn}} = a_v^b \cdot \omega \sum\nolimits_{z_b=1}^{Z_b} \mathbb{E}_{\mathbf{h}} \bigl[ \log_2 (1 + \Gamma_{b,z}^{v, \text{dn}} ) \bigr],  
\end{equation}
where $\mathbb{E}_{\mathbf{h}}[\cdot]$ is the expectation over $\mathbf{h}_{b,z}^v$.

\subsubsection{Uplink Communication Model}
For the uplink, assuming linear receiver vector $\mathbf{g}_{v,z}^{b} \in \mathbb{C}^{1 \times n_b}$, we calculate the effective received signal at RSU $b$ from vehicle $v$ as 
\begin{equation}
\label{uplink_rx_sig_at_b}
    y_{v,z}^{b,\text{up}} = \sqrt{P_{v}^b} a_v^b \mathbf{h}_{v,z}^{b} ({\mathbf{g}_{v,z}^{b}})^H s_{v}^b +\eta,
\end{equation}
where $P_{v}^b$ is the uplink transmission power of vehicle $v$ and $s_v^b$ is the intended uplink transmitted symbol of $v$.
Moreover, this gives us the following uplink SNR
\begin{equation}
\label{uplink_snr}
    \Gamma_{b,z}^{v,\text{up}} = {P_v^b a_v^b \bigl | \mathbf{h}_{v,z}^{b} ({\mathbf{g}_{v,z}^{b}})^H  \bigr |^2} /{\omega \sigma^2}.
\end{equation}
Similar to $C_v^{\text{dn}}$, we calculate the uplink rate as 
\begin{equation}
\label{uplink_rate}
    C_v^{\text{up}} = a_v^b \cdot \omega \sum\nolimits_{z_b=1}^{Z_b} \mathbb{I}_{b,z}^v \cdot \mathbb{E}_{\mathbf{h}} \bigl [ \log_2 ( 1 + \Gamma_{b, z}^{v,\text{up}} ) \bigr],
\end{equation}
where $\mathbb{I}_{b,z}^v \in \{0,1\}$ is an indicator function that takes value $1$ when $z_b \in \mathcal{Z}_b$ is assigned to vehicle $v$.

\section{Communication and Computation Aware Vehicular FL Problem}
\label{Dealay_FedProx_Sec}

We propose vehicular FL by incorporating heterogeneity in mobility, communication and computation.

\subsection{Vehicular FL}
We assume that each vehicle can use its on-board sensors to collect data, such as longitude, latitude, velocity, acceleration, weather information, etc., that can be used for model training.
Without loss of generality, denote the data sensing interval of the vehicles by $\Delta t$.
Denote vehicle $v$'s dataset at time $t$ by $\mathcal{D}_v^t = \{ \mathbf{x}_{v}^i, y_v^i\}_{i=1}^{t}$, where $\mathbf{x}_v^i$ and $y_v^i$ are the $i^{\text{th}}$ feature set and corresponding label, respectively.
The entire dataset available at time $t$ is denoted as $\mathcal{D}^t=\bigcup_{v=1}^V \mathcal{D}_v^t$.
As the central server does not have access to dataset $\mathcal{D}^t$, it aims to solve following optimization problem \cite{mcmahan2017communication}:
\begin{equation}
\label{serverObjective}
\begin{aligned}
	\underset{\mathbf{w}}{\text{minimize }} &f(\mathbf
	w) = \mathbb{E}_v \left[ f_v(\mathbf{w})\right] = \sum\nolimits_{v=1}^V p_v \cdot f_v(\mathbf{w}),\\
	\text{s. t.} \quad & \quad \mathbf{w}^{(1)}=\mathbf{w}^{(2)}, \dots, = \mathbf{w}^{(V)}=\mathbf{w},
\end{aligned}
\end{equation}
where $\mathbf{w}$ is the global model parameters, $f_v(\mathbf{w}) \coloneqq \mathbb{E}_{\{\mathbf{x}_v^i, y_v^i\} \sim \mathcal{D}_v^t} \left[ f_v(\{\mathbf{x}_v^i, y_v^i\}, \mathbf{w})\right]$ is the local empirical risk function for agent $v$ and $\sum_{v=1}^V p_v=1$ 
with the probability $p_k = {|\mathcal{D}_v^t|}/{|\mathcal{D}^t|}$.

This FedAvg based FL works well when the agents have IID data distribution and homogeneous computation power.  
In our case, the vehicle agents have (a) diverse on-board sensors that lead to non-IID data distribution and (b) heterogeneous on-board computation power.
Therefore, FedProx based FL \cite{MLSYS2020_38af8613} is more suitable because it is designed for agents with different computation resources. 
Accordingly, FedProx accepts partial works of the stragglers while each agent tends to solve its local
optimization problem inexactly.

We extend the FedProx based FL to the Internet of Vehicles. 
At the beginning of a global round denoted by $k$, the server uniformly selects a set of agents from the vehicle pool.
Denote the selected agent set by $\mathcal{V}_k \subseteq \mathcal{V}$.
Each agent $v \in \mathcal{V}_k$ then receives the global model $\mathbf{w}_k$ and perform local model training to minimize the following objective function
\begin{equation}
\label{agent_Objective_Fun_FedProx}
\begin{aligned}
    \breve{f}_v(\mathbf{w}; \mathbf{w}_k) = f_v (\mathbf{w}) + (\mu/2) \left\Vert \mathbf{w} - \mathbf{w}_k \right\Vert^2, 
\end{aligned}
\end{equation}
where the proximal term is added to control heterogeneity and $\mu \ge 0$ is the penalty parameter.

Agent $v\in \mathcal{V}_k$ solves problem (\ref{agent_Objective_Fun_FedProx}) $\gamma_v^k$-inexactly for solution $\mathbf{w}^{(v,*)}_k$ such that $\left\Vert \nabla \breve{f}_v(\mathbf{w}^{(v,*)}_k; \mathbf{w}_k) \right\Vert \leq \gamma_v^k \left\Vert \nabla \breve{f}_v(\mathbf{w}_k; \mathbf{w}_k) \right\Vert$ \cite{MLSYS2020_38af8613}.
The parameter $\gamma_v^k$ defines how much local
computation to be performed by agent $v$ for solving its respective local sub-problem.

Upon receiving $\mathbf{w}^{(v,*)}_k$s from all learning agents, the central server averages these model parameters to obtain $\mathbf{w}_{k+1}$.
We will explicitly introduce communication and computation aware FL solutions in Section \ref{Sec_Soln_Fedprox}.

\subsection{Delays in Vehicular FL}
We consider three main delays between two global communication rounds as described in the following.

\subsubsection{Queuing Delay}
Queuing delay is the waiting time of a vehicular agent before being scheduled by the associated RSU. It can be an important delay contributor in wireless networks.
Note that we only consider uplink queuing delay as the RSUs multicast in the downlink.
Denote the uplink queuing delay of agent $v$ by $d_v^{q, \text{up}}$, which is the time difference from the time agent $v$ finishes local model training to the time agent $v$ is scheduled to offload the trained model.

\subsubsection{Local Model Training Delay}
Recall that the vehicles have heterogeneous on-board processing unit power.
Denote agent $v$'s processing power by $\rho_v$ cycles per second.
If per sample data requires $\eta_v$ cycles for processing, then the delay for one iteration model training is $d_v^{\text{cmp},l} = \left[\left(\eta_v \left\vert\mathcal{D}_v\right\vert\right)\right]/\rho_v$.

\subsubsection{Communication Delay}
With vehicular agents, we consider delay from the wireless link.
We calculate the downlink and uplink delays based on downlink rate $C_v^{\text{dn}}$, uplink rate $C_v^{\text{up}}$ and payload size.
Assuming the FL model parameter is $d$-dimensional, we calculate the required number of bits as $S = \sum_{i=1}^{d}FPP_i$, where $FPP_i$ represents the floating point precision for element $i$.
As such, at the beginning of a global round,  
we calculate the model distribution delay to agent $v$ via RSU-vehicle downlink communication as follows:
\begin{equation}
\label{downlink_delay}
    d_{v}^{\text{down}} = \kappa \times \text{min} \{T : \kappa \times \bigl(\sum\nolimits_{\bar{t}=1}^T C_v^{\text{dn}}(\bar{t})\bigr) \geq S, T \in \mathbb{Z}^{+}\}, 
\end{equation}
where $\kappa$ is the transmission time interval (TTI) and $C_v^{\text{dn}}(\bar{t})$ is the achievable downlink capacity based on the channel realization at slot $\bar{t}$. 

Similarly, we can calculate the delay to offload agent $v$'s trained model as follows:
\begin{equation}
\label{uplink_delay}
    d_{v}^{\text{up}} = \kappa \times \text{min} \{T : \kappa \times \bigl(\sum\nolimits_{\bar{t}=1}^T C_v^{\text{up}}(\bar{t})\bigr) \geq S, T \in \mathbb{Z}^{+}\},
\end{equation}
where $C_v^{\text{up}}(\bar{t})$ is the achievable uplink capacity based on the channel realization at slot $\bar{t}$. 

\subsubsection{Total Delay in Each Global Communication Round}
To this end, we calculate the cumulative delays between two global rounds. The total delay is the summation of the global model distribution delay, local model training delay, uplink queuing delay and local model offloading delay of the agents.
We express the total delay for agent $v$ as follows:
\begin{equation}
\label{total_delay_calculation}
    \mathrm{d}_v^{\text{tot}} = d_{v}^{\text{down}} + d_v^{\text{cmp}} + d_v^{q, \text{up}} + d_{v}^{\text{up}}.
\end{equation}
In this work, we consider a synchronous agent-server update procedure as described in Section \ref{threshold}. 
The global round update clock time is fixed and known to the agents.
More specifically, the central server sets a hard threshold denoted as $\mathrm{d}^{\text{thr}}$, by which the server needs to distribute the global model, agents need to locally train model and offload the trained models back to the server.
In other words, $\mathrm{d}_v^{\text{tot}} \le \mathrm{d}^{\text{thr}}$ must hold for this synchronous update process.
Clearly, the underlying wireless communication and computation aspects affect whether the agents can successfully contribute to the learning process or not.

\section{Proposed Mobility, Communication and Computation Aware FL Solution}
\label{Sec_Soln_Fedprox}

At a high level, our goal is to optimize the IoV network so that $\mathrm{d}_v^{\text{tot}} \leq \mathrm{d}^{\text{thr}}$.
In order to do this, we first optimize the model distribution delay in the downlink.

\subsection{Model Distribution Delay Optimization}
To optimize the downlink model distribution delay, each RSU aims to maximize the minimum data rate for all of its associated agents. 
As such, each RSU finds the downlink multicasting beamforming vector $\mathbf{g}_b^z$ for all pRBs by solving the following optimization problem:
\begin{equation}
\label{donwlinkMulticastBeamforming0}
\begin{aligned}
    \underset{\mathbf{g}_b^z, ~ \forall z_b \in \mathcal{Z}_b} {\text{ maximize }} \underset{\forall v \in \mathcal{V}_k^b} { \text{ min }} & \quad \left | ({\mathbf{h}_{b,z}^v})^H \mathbf{g}_{b}^z \right |^2, \\
    \text{subject to:} \qquad & \left\Vert \mathbf{g}_{b}^z \right\Vert^2 \leq 1,
\end{aligned}
\end{equation}
where $\mathcal{V}_k^b \subseteq \mathcal{V}_k$ is set of agents associated with RSU $b$ in global round $k$.

This is a classical multicasting beamforming problem.
Note that $\left | ({\mathbf{h}_{b,z}^v})^H \mathbf{g}_{b}^z \right |^2 = \Tr \left(\mathbf{g}_{b}^z {\mathbf{g}_{b}^z}^H \mathbf{h}_{b,z}^v ({\mathbf{h}_{b,z}^v})^H \right)$.
Denote $\mathbf{H}_{b,z}^v = \mathbf{h}_{b,z}^v ({\mathbf{h}_{b,z}^v})^H$ and $\mathbf{G}_{b,z}=\mathbf{g}_{b}^z ({\mathbf{g}_{b}^z})^H$.
Then, problem (\ref{donwlinkMulticastBeamforming0}) can be reformulated as follows:
\begin{equation}
\label{donwlinkMulticastBeamforming1}
\begin{aligned}
    &\underset{\mathbf{G}_{b,z}, ~ \forall z_b \in \mathcal{Z}_b} {\text{ maximize }} \quad \underset{\forall v \in \mathcal{V}_k^b}  { \text{ min }}  \quad \Tr \left(\mathbf{G}_{b,z} \mathbf{H}_{b,z}^v\right), \\
    &\text{subject to:} \quad  \Tr \left( \mathbf{G}_{b,z} \right)=1, \mathbf{G}_{b,z} \succeq 0, \mathrm{rank} (\mathbf{G}_{b,z})=1.
\end{aligned}
\end{equation}
Note that (\ref{donwlinkMulticastBeamforming1}) is non-convex due to the $\mathrm{rank} (\mathbf{G}_{b,z})=1$ constraint. 
We can relax this constraint to obtain the following relaxed convex problem. 
\begin{equation}
\label{donwlinkMulticastBeamforming2}
\begin{aligned}
    \underset{\mathbf{G}_{b,z}, ~ \forall z_b \in \mathcal{Z}_b} {\text{ maximize }} \quad \underset{\forall v \in \mathcal{V}_k^b} { \text{ min }} & \quad \Tr \left(\mathbf{G}_{b,z} \mathbf{H}_{b,z}^v\right), \\
    \text{subject to:} \qquad & \Tr \left( \mathbf{G}_{b,z} \right)=1, \mathbf{G}_{b,z} \succeq 0.
\end{aligned}
\end{equation}
Optimization problem (\ref{donwlinkMulticastBeamforming2}) is in the well-known semidefinite problem (SDP) form.
Each RSU can solve this downlink multicasting beamforming optimization problem using widely popular convex optimization solver such as CVX \cite{grant2009cvx}.
Each RSU finds the downlink multicasting beamforming vector and distributes the model to all associated agents. 
Note that since the entire bandwidth is used for this downlink distribution, the $d_{v}^{\text{down}}$ is relatively low. 
Moreover, the RSU-agent associations will remain unchanged for this short time\footnote{ 
For example, if RSU coverage radius is $300$ meters and agent velocity is $120$ miles/hour, the agent is expected to be within the RSU's coverage for $\approx 11.18$ seconds. 
In wireless networks, usually TTI $\kappa \leq 1$ millisecond.}.

\subsection{Local Model Training}
\label{threshold}
This work considers a complete synchronous learning framework, where the server provides all agents a deadline $\mathrm{d}_{\text{k}}^{\text{cmp}}$ for their local model training. 
In other words, during round $k$, the agent receives the model and performs local model training until $\mathrm{d}_{\text{k}}^{\text{cmp}}$ expires.
Recall that the FL global round update clock time is known to all agents.
Therefore, during global round $k$, upon receiving the global model $\mathbf{w}_k$, each agent $v \in \mathcal{V}_k$ can determine the remaining time budget for its local model computation as
\begin{equation}
    d_v^{\text{cmp}} =  \mathrm{d}_{\text{k}}^{\text{cmp}} - d_{v}^{\text{down}}.
\end{equation}
Therefore, agent $v$ determines its local model training iterations as 
\begin{equation}
\label{localIterDeter}
    L_v = \left \lfloor d_v^{\text{cmp}}/ d_v^{\text{cmp},l} \right \rfloor,
\end{equation} 
where $\lfloor \cdot \rfloor$ is the floor operation.
This essentially means that agent $v\in\mathcal{V}_k$ performs $L_v$ local stochastic gradient decent (SGD) steps to minimize its local objective function defined in (\ref{agent_Objective_Fun_FedProx}).
Please note that, unlike FedAvg that considers equal $L_1=\dots=L_V$ \cite{mcmahan2017communication}, FedProx allows heterogeneous device participation to utilize agents' resources efficiently \cite{MLSYS2020_38af8613}. 

Note that FedAvg is a special case of the proposed FedProx based solution if the server sets an common training iteration $L$ and $\mu$ is set to $0$ in problem (\ref{agent_Objective_Fun_FedProx}).

\subsection{Local Model Offloading Optimization}
Upon finishing the local model training, the agent requests uplink radio resources from its associated RSU to offload the trained model $\mathbf{w}_k^{v}$. 
The RSU then allocates the pRB for this uplink communication.
Assume RSUs have perfect CSI. 
Therefore, each RSU can use maximal ratio combining (MRC) to model the receiver beamforming vector, i.e., $\mathbf{g}_{b,z}^v=\breve{\mathbf{h}}_{b,z}^v/ \| \breve{\mathbf{h}}_{b,z}^v \|$.
Moreover, depending on the pRB allocation, the uplink queuing delay $d_{v}^{\text{up}}$ is known to the associated RSU. 
In this work, we apply round-robin scheduling for simplicity.
To that end, each RSU allocates its pRBs to the scheduled agents to maximize the network's uplink throughput.
In other words, each RSU aims to solve following optimization problem:
\begin{equation}
\label{uplinkSR}
\begin{aligned}
    \underset{\mathbb{I}_{b,z}^v, ~ \forall v \in \mathcal{V}_k^{b}} {\text{ maximize }} \quad & \omega \sum\nolimits_{z_b=1}^{Z_b} \mathbb{I}_{b,z}^v \cdot  \log_2 \left( 1 + \Gamma_{b, z}^{v,\text{up}} \right),\\
    \text{subject to } \quad & \sum\nolimits_{z_b=1}^{\mathcal{Z}_b} \mathbb{I}_{b,z}^v = 1, \sum\nolimits_{v \in \mathcal{V}_k^{b}} \mathbb{I}_{b,z}^v = 1, \\
    & \sum\nolimits_{z_b=1}^{\mathcal{Z}_b} \sum\nolimits_{v \in  \mathcal{V}_k^{b}} \mathbb{I}_{b,z}^v = |\mathcal{Z}_b|, 
\end{aligned}
\end{equation}
where the first constraint is to allocate only one pRB to each scheduled agent, while the second constraint is adopted to assign a pRB to only one agent.
Moreover, the last constraint ensures that all pRBs are allocated.
Note that while $|\mathcal{V}_k|$ can be greater than $|\mathcal{Z}_b|$, the RSU can only schedule $|\mathcal{Z}_b|$ agents in a each scheduling $\kappa$, i.e., $|\mathcal{V}_k^{b}| = |\mathcal{Z}_b|$.

Problem (\ref{uplinkSR}) is a mixed combinatorial optimization problem and NP-hard. 
To solve this problem, we stack the SNRs over all pRBs into a gain matrix $\mathbf{G}_b$ and use the well-known Hungarian algorithm \cite{kuhn1955hungarian}.
Note that this algorithm guarantees optimal pRB allocation with time complexity of $\mathcal{O} (|\mathcal{Z}_b|^3)$ \cite{kuhn1955hungarian}.
Algorithm \ref{pRB_Allocation_Alg} summarizes the process.

\SetInd{0.8em}{0.2em}
\HinTwoColAlg
\begin{algorithm} [t!]
\small
\caption{Optimal Radio Resource Allocation}
\label{pRB_Allocation_Alg}
\SetAlgoLined
\textbf{Input}: $\{\Gamma_{b,z}^{v,up}\}$ for all active RSUs and vehicle agents \;
\For {$b \in \mathcal{B}$} {
    Schedule agents following round-robin scheduling\; 
    \uIf{$ \bigl \vert \mathcal{V}_k^{b} \bigr \vert  \neq 0$} {
        Initiate gain matrix, $\mathbf{G}_b = zeros(Z_b \times \bigl \vert  \mathcal{V}_k^{b}\bigr \vert )$ \;
        \For{$v \in \mathcal{V}_k^{b}$} {
            \For{$z_b \in \mathcal{Z}_b$}{
                $\mathbf{G}_b[z_b, v] \gets \Gamma_{b,z}^{v,\text{up}}$\;
            }
        }
        Use $\mathrm{Hugarian}$ algorithm \cite{kuhn1955hungarian} to get optimal $\mathbb{I}_{b,z}^v $s \;
    }
}
\end{algorithm} 
\begin{algorithm} [t!]
\small
\caption{Vehicular FL Algorithm}
\label{FedPRox_VL_Alg}
\SetAlgoLined
\textbf{Input}: initial global model $\mathbf{w}$\;
\For{all global communication round, $k \in K$} {
    \tcc{\textit{\textbf{At the beginning of communication round} $k$}}
    Server selects learning agents $\mathcal{V}_k \subseteq \mathcal{V}$ \;
    Each RSU solves optimization problem (\ref{donwlinkMulticastBeamforming2}) and distributes} the global model $\mathbf{w}_k$ \Comment{\textit{Central server broadcasts global model $\mathbf{w}_k$, agents set $\mathcal{V}_k$, $\mathrm{d}_{\text{k}}^{\text{cmp}}$ and threshold $\mathrm{d}^{thr}$ to RSUs. RSUs then forward the global model $\mathbf{w}_k$ and the updated time information $d_v^{\text{cmp}}$ to their associated $v \in \mathcal{V}_k$}}\;
    \tcc{\textit{\textbf{Local model training}}}
    \For{all $v \in \mathcal{V}_k$ in parallel} {
        Prepare the local training dataset $\mathcal{D}_{v}^{(k)}$\;
        Determine local training iterations $L_v$ using (\ref{localIterDeter})}\;
        Perform local training to minimize loss function defined in (\ref{agent_Objective_Fun_FedProx})\;
    \tcc{\textit{\textbf{At the end of global round} $k$} }
    All $v \in \mathcal{V}_k$ request uplink radio resources from their associated RSUs to offload their locally trained model $\mathbf{w}_k^{v}$ \;
    Each RSU $b$ solves optimization problem (\ref{uplinkSR}) to get the optimal pRB allocation and schedules its associated agents $v \in \mathcal{V}_k^b$ in round-robin manner\;
    Each agent $v \in \mathcal{V}_k$ then offloads $\mathbf{w}_k^{v}$ to the central server via the associated RSU $b$\;
    The central server performs model aggregation to update global model using: 
         $\mathbf{w}_{k+1} \gets (1/\left\vert\mathcal{V}_k\right\vert) \sum\nolimits_{v=1}^{\left\vert\mathcal{V}_k\right\vert} \mathbf{w}_k^{v}$.
\textbf{Output}: trained global model $\mathbf{w}^{*}$
\end{algorithm}

\subsection{Summary of the Proposed FL Solution}
The proposed mobility, communication and computation aware FL solution is summarized in Algorithm \ref{FedPRox_VL_Alg}. In each communication round, server selects agents to train the model. Server broadcasts the model $\mathbf{w}_k$, the agent set $\mathcal{V}_k$, time deadlines $\mathrm{d}_k^{\text{cmp}}$ and $\mathrm{d}^{\text{thr}}$ to RSUs. Each RSU then computes optimal communication delays for its associated agents and broadcasts the model and the updated time information to its agents. Based on the updated time information $d_v^{\text{cmp}}$ and its local dataset $\mathcal{D}_k^{v}$, each agent performs independent model training for $L_v$ iterations. At training deadline $\mathrm{d}_k^{\text{cmp}}$, each RSU schedules its agents to offload their local models to satisfy deadline threshold $\mathrm{d}^{\text{thr}}$.

\section{Performance Evaluation}
\label{Sec_Simulation}
In this section, we first describe our simulation setup and then present simulation results and discussions.

\subsection{Simulation Setting and Dataset Description}
We use real vehicle traces in the Mobile Century Dataset collected by the Mobile Millennium project in $2008$ \cite{herrera2010evaluation}. 
This dataset contains data collected for $77$ vehicles between 10:00am and 18:00pm on Interstate $880$ (I-$880$), CA, USA and has vehicle trajectory traces with an approximate interval of 3 seconds.
We use GPS log datasets that have four features in each data sample: $\{unixtime, ~latitude, ~longitude, ~speed\}$.
From GPS log data of these $77$ vehicles, we choose $61$ datasets after initial screening.
Note that the traces contain approximately $20.1$ miles stretch of the I-$880$. 
To serve these vehicles, we place RSUs along the I-$880$ for the entire stretch. 
Particularly, we assume each RSU has a coverage radius of $500$ meters.
Moreover, the same frequencies are reused after $\approx 2000$ meters to handle interference issues.

For our learning model, we use $1$ gated recurrent unit (GRU) layer with $16$ hidden layers, SGD (momentum=0.95) optimizer with learning rate $0.001$, $\mu=0.1$, $l=15$, and $h \in \{1, \dots, 12\}$.
We select $10$ agents with $\rho_v \in Uniform(0.2, 0.8)$ GHz and $\eta_v \in Uniform(10^4, 40^4)$ cycles/sample.
Furthermore, the agents use $11:00-16:30$ trajectory data samples as their training dataset. We devise the global round update time in such a way that at the beginning, there are enough data samples to train the model, and the training data samples $\mathcal{D}_k^{v}$ only use data samples from $11:00:00$ to the global round update time in a causal manner.
This means that the proposed FL solution is implemented in an online fashion, in which the vehicles gather data as they travel along their trajectory.
Other key V2X simulation parameters are listed in Table \ref{V2X_Sim_Param}.

\begin{table}[!t]
\fontsize{8}{8}\selectfont
\caption{\textbf{V2I/I2V: Key Simulation Parameters}} 
\begin{tabular}{|K{4cm}|K{3cm}|}
    \hline
    \textbf{Description} & \textbf{Value} \\ \hline
    Antenna per RSU $n_b$ & $4$ \\ \hline
    pRB per RSU $Z_b$ & $10$ \\ \hline
    \{$v, b$\} antenna heights & \{$3$, $25$\} m \\ \hline
    \{$v, b$\} antenna gains & \{$3, 8$\} dBi \\ \hline
    \{$v, b$\} receiver noise figures & \{$9, 5$\} dB \\ \hline
    Noise power $\sigma^2$   & -$174$ dBm \\ \hline
    $P_b^{max}, P_v^b$ & $30, 23$ [dBm] \\ \hline
    TTI $\kappa$ & $1$ ms \\ \hline
    Carrier frequency & $2.4$ GHz \\ \hline
    pRB bandwidth $\omega$ & $180$ KHz \\ \hline
    $\Omega_b, \Omega$ & $1.8, 3.6$ [MHz] \\ \hline 
    Frequency reuse & $\approx 2000 $ m \\ \hline
\end{tabular}
\label{V2X_Sim_Param}
\end{table}

\subsection{FL Task : Vehicle-Specific Power (VSP) Prediction}
While the proposed vehicular FL platform is not task-dependent, we consider a multi-horizon vehicle-specific power (VSP) prediction task to validate our vehicular FL solutions. In essence, VSP provides the estimation of the required power demand for the vehicle.
For zero road grade, the VSP can be calculated as \cite{koupal2005moves2004}
\begin{equation}
\label{VSP_Eqn}
    P_v^t = \left(\frac{c_1}{c_2}\right) \frac{A u_v^t}{m_v} + \left(\frac{c_1^2}{c_2}\right) \frac{B {u_v^t}^2}{m_v} + \left( \frac{c_1^3}{c_2} \right) \frac{C {u_v^t}^3}{m_v} + c_1^2 u_v^t a_v^t,
\end{equation}
where $m_v$, $u_v^t$ and $a_v^t$ are vehicle $v$'s weight, instantaneous velocity and instantaneous acceleration, respectively.
Moreover, $A$, $B$, $C$, $c_1$ and $c_2$ are coefficients described in \cite{koupal2005moves2004}.

\begin{figure}[t!]
\begin{minipage}{0.24\textwidth}
    \includegraphics[width=\textwidth]{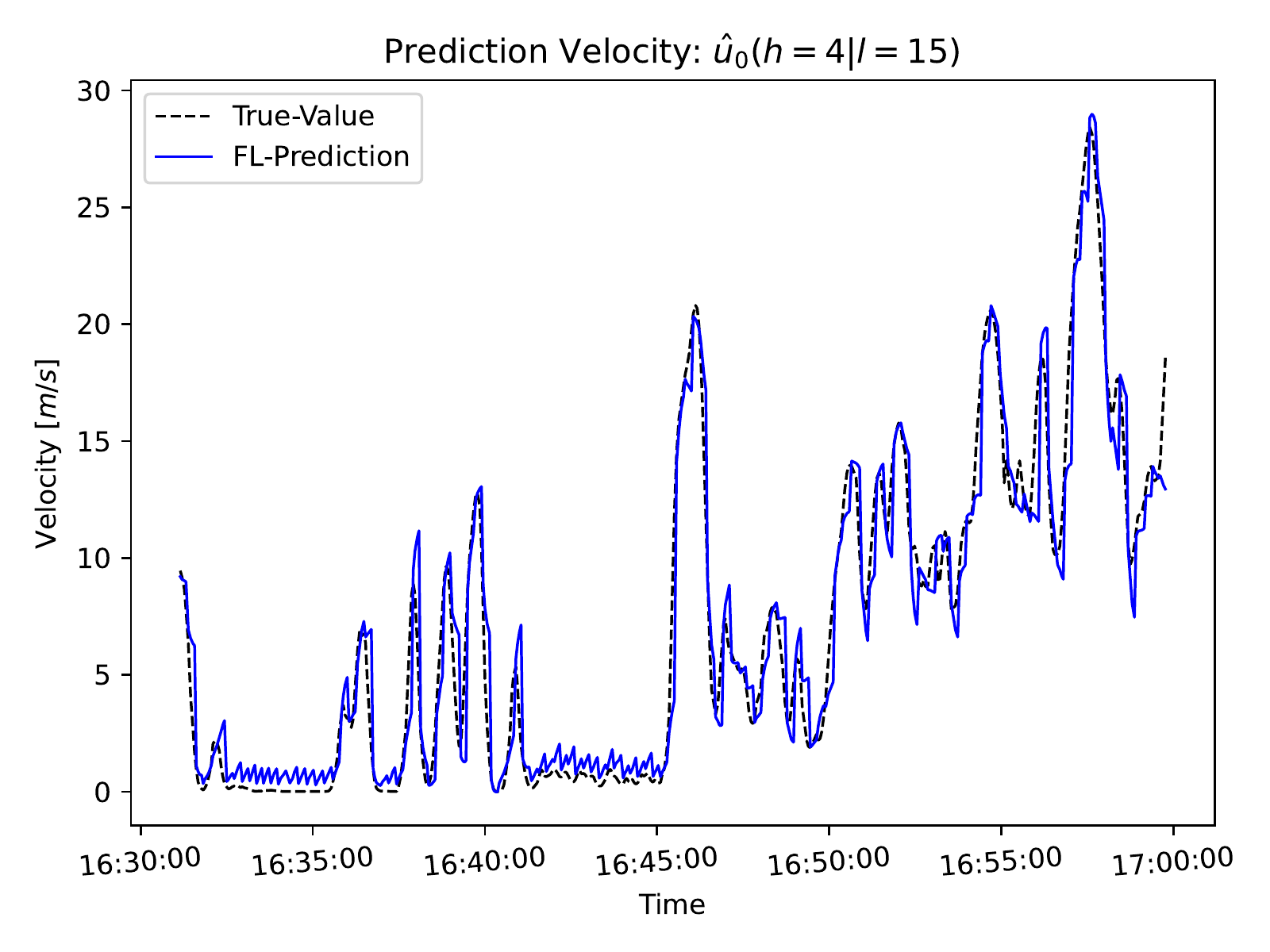} 
    \caption{Velocity prediction for vehicle $0$}
    \label{velo_comparison}    
\end{minipage} \hspace{0.001\textwidth}
\begin{minipage}{0.235 \textwidth}
    \includegraphics[width=\textwidth]{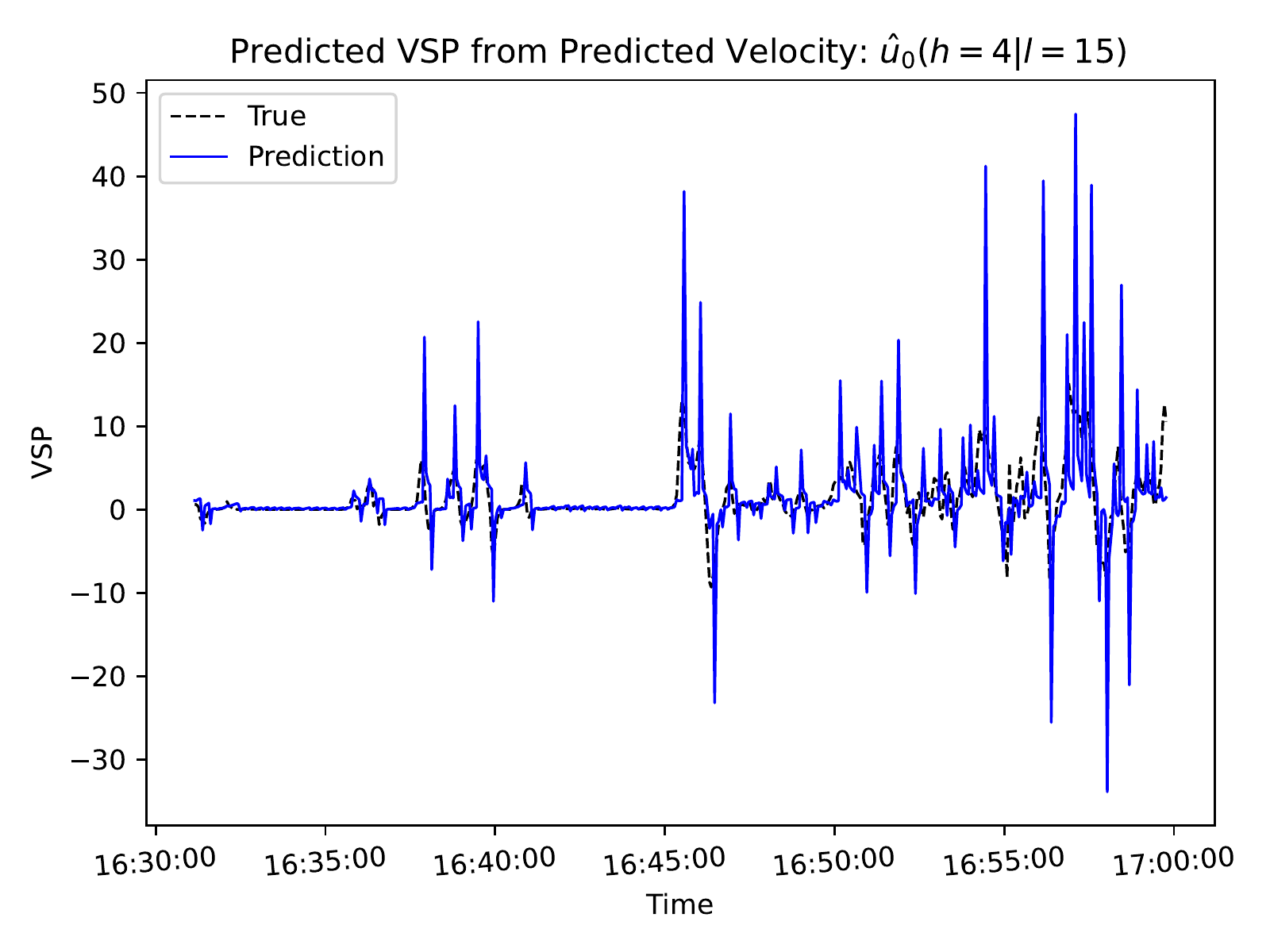} 
    \caption{VSP calculation from predicted velocity for vehicle $0$}
    \label{VSP_Veh_0_forecast_4}
\end{minipage}
\end{figure}

The VSP can be  predicted in two ways: (i) predicting velocity and then using the predicted velocity to compute VSP via Eq. (\ref{VSP_Eqn}) and (ii) predicting VSP directly. The approach (i) does not work well because VSP is expressed as a third-order polynomial in velocity with additional velocity acceleration product term. Accordingly, it is very sensitive to velocity variation. Even with near ground truth velocity prediction as shown in Fig. \ref{velo_comparison}, the VSP accuracy can be poor as shown in Fig. \ref{VSP_Veh_0_forecast_4}. 
Therefore, we apply approach (ii) in this work. Note that in order to predict VSP directly with approach (ii), we need a suitable dataset. However, the dataset \cite{herrera2010evaluation} does not have VSP information. 
As such, we use (\ref{VSP_Eqn}) to calculate VSP and then use the calculated VSP for model training and evaluation in what follows.

\subsection{Simulation Results and Discussions}
We evaluate the proposed FL solutions in four aspects: (i) delay, (ii) prediction accuracy, (iii) robustness, and (iv) prediction horizon. It is worth noting that each agent $v$ randomly samples $50\%$ data samples from $\mathcal{D}_k^{v}$ for its local model training\footnote{The percentage of data samples can be adjusted to meet the deadlines.}. 
Moreover, we choose $\mathrm{d}^{\text{thr}}=3$ seconds and accordingly set estimated local iterations $L_v$ for all $v \in \mathcal{V}_k$ in each communication round $k$. 
\subsubsection{Delay in Distributed Model Training}
We first show delays between two global communication rounds using the simulation parameters in Table \ref{V2X_Sim_Param}. 

\begin{figure}[t!]
\centering
\includegraphics[trim=0.25cm 0.25cm 0.25cm 0.25cm, clip=true,width=0.48\textwidth]{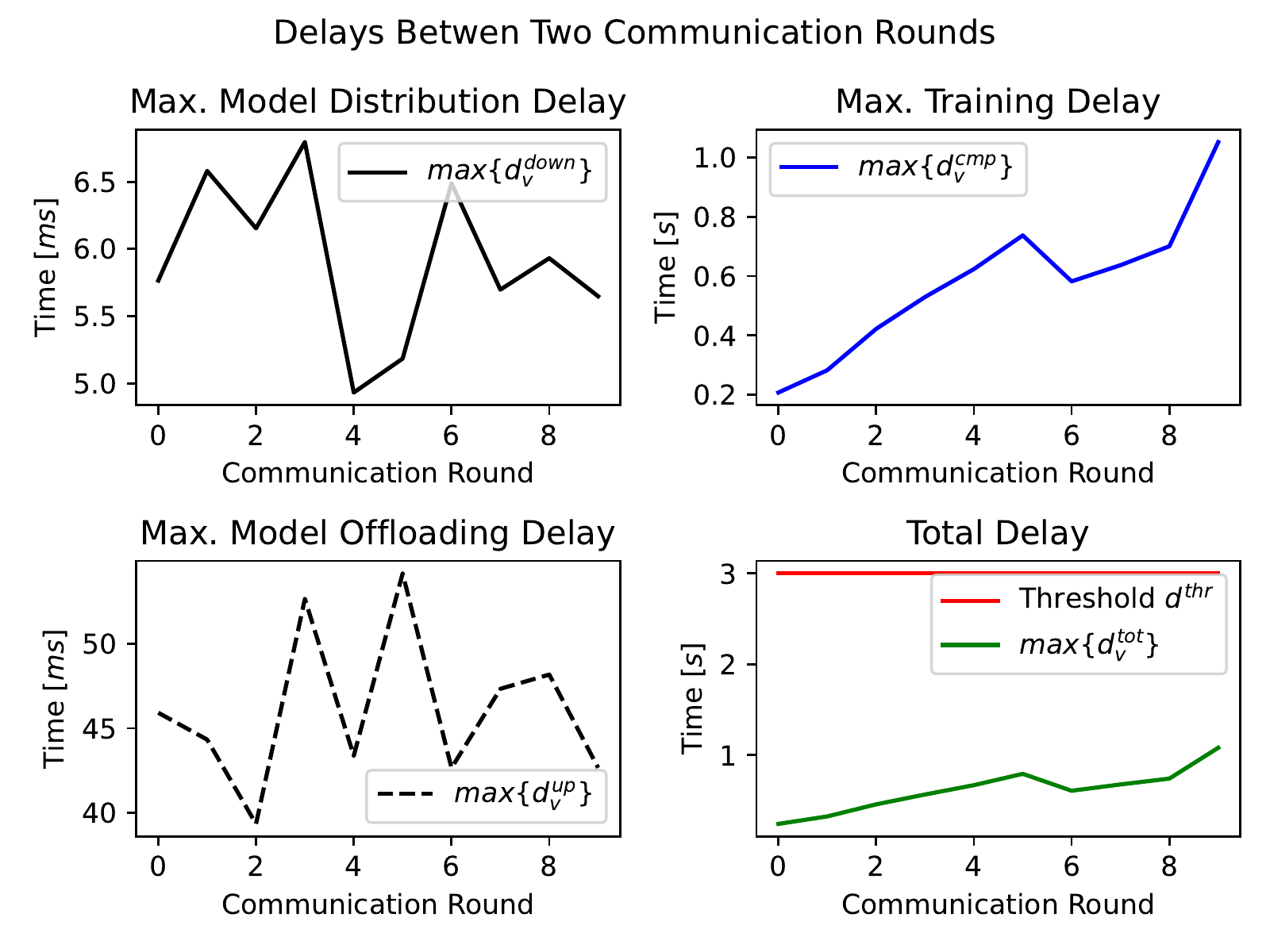} 
\caption{Delay between two communication rounds}
\label{total_delay}
\end{figure}
\begin{figure}[!t]
    \centering
    \includegraphics[width=0.5\textwidth]{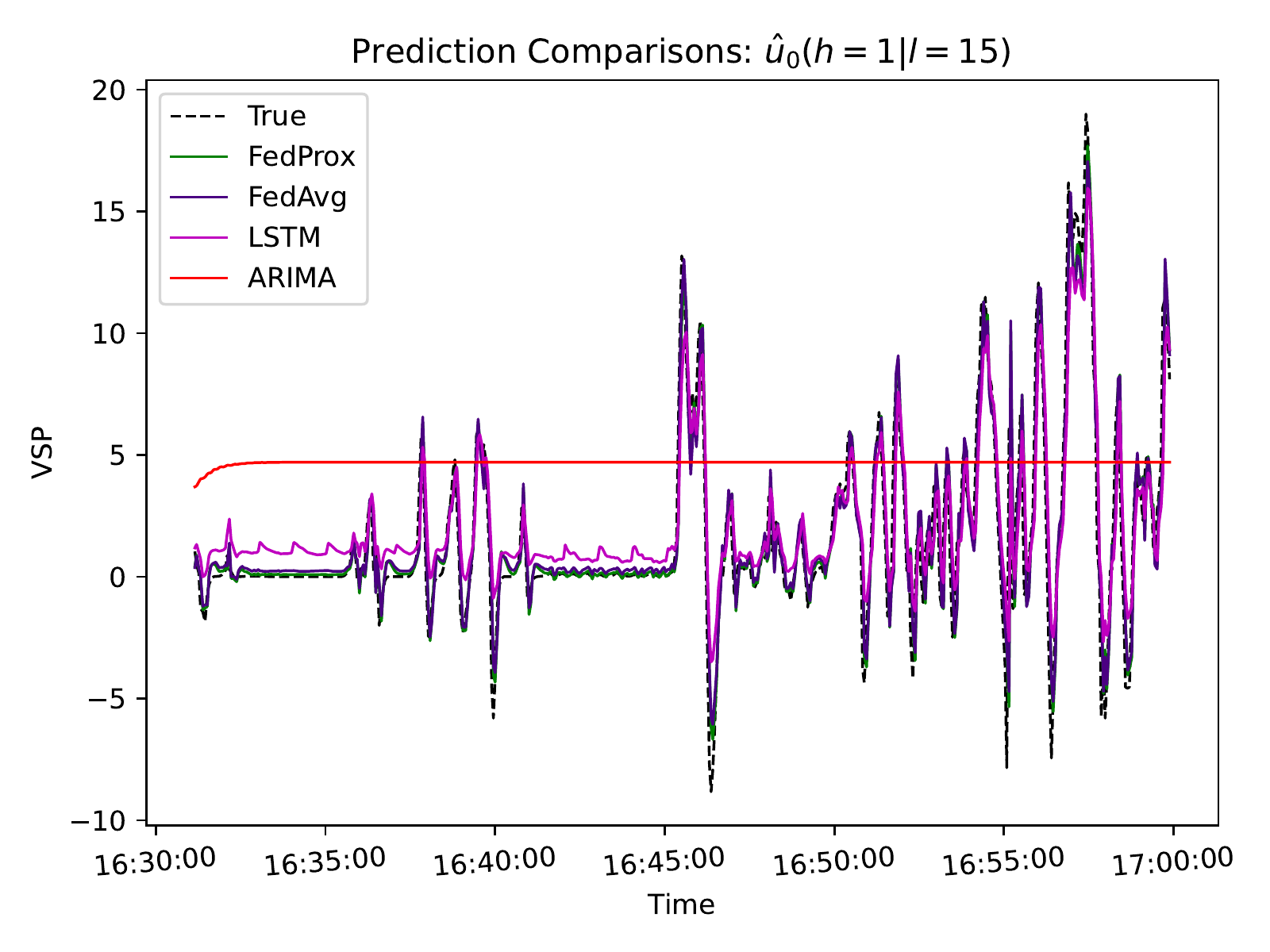} 
    \caption{Predicted VSP with different machine learning models}
    \label{baselineComparison}
\end{figure}

Given a model training time threshold $\mathrm{d}^{\text{thr}}$, the RSUs optimizes the multicast beamforming vector to maximize the minimum downlink data rate at the agents by solving optimization problem (\ref{donwlinkMulticastBeamforming2}).
As expected, the downlink model distribution delay $d_{v}^{\text{down}}$ is significantly lower. As shown in the first subplot of Fig. \ref{total_delay}, the maximum distribution time is at most $7 \times \kappa$, i.e., $7$ ms.

However, the uplink model offloading time $d_{v}^{\text{up}}$ is significantly longer than $d_{v}^{\text{down}}$ as shown in the third subplot of Fig. \ref{total_delay}. 
In fact, it requires more TTIs to offload local models because agents have limited radio resources and battery power.
This is also observed in our simulation.
More specifically, we observe that it requires $\approx 55$ TTIs to offload an agent's model in the uplink.
This means that we cannot simply consider fixed channel $\mathbf{h}_{v,z}^b$ conditions to calculate $d_{v}^{\text{up}}$.
Therefore, solving optimal pRB allocation problem (\ref{uplinkSR}) is necessary.

Since the on-board computation power of the agents is limited, the dominant factor in total delay is the local model training delay\footnote{If the model size is significantly large, the communication overhead may become higher.} as shown in the second subplot of Fig. \ref{total_delay}.
As the global round increases, agents collect more data in their datasets $\mathcal{D}_k^{v}$s. 
Therefore, the local training time increases as the global round increases. As a result,  total delay increase as well as shown in the fourth subplot of Fig. \ref{total_delay}.

Note that, given a time threshold $\mathrm{d}^{\text{thr}}$, the choice of $L_v$ and $\mathrm{d}_{\text{k}}^{\text{cmp}}$ are critical to guarantee $\mathrm{d}_v^{\text{tot}} \leq \mathrm{d}^{\text{thr}}$.
Particularly, to choose $\mathrm{d}_{\text{k}}^{\text{cmp}}$, the server should consider the impact of the uncertain channel condition, available radio resources of the RSU and corresponding expected queuing delay.

\subsubsection{VSP Prediction Accuracy}
In this subsection, we show the performance of different ML models on VSP prediction task. 
We compare our proposed FedProx based FL model with FedAvg based FL model + our communication framework, independent LSTM model and a widely used statistical time-series baseline, namely the AutoRegressive Integrated Moving Average (ARIMA) model.
Particularly, the statistical ARIMA works well when the data samples are homogeneous and have a clear trend.
However, our data are essentially the behavior of drivers, i.e., highly heterogeneous.
As such, it is hard to find a clear trend. 
Therefore, ARIMA is expected to perform poorly.
On the other hand, the independent LSTM model only uses agent's own data for training, which works for the homogeneous data distribution case. 
However, in the highly heterogeneous data distribution case, the prediction accuracy of the LSTM model is expected to degrade.
Our proposed FedProx based FL model is exposed to diverse training samples of the selected agents.
It is thus anticipated to perform better than these baseline models.
This tendency is validated in our simulation. 
Fig. \ref{baselineComparison} takes vehicle $0$ as an example. 
Clearly, our FedProx based FL model can deliver near-ground truth VSP prediction, whereas the independent LSTM model and ARIMA model fail to capture the temporal dynamics. Furthermore our FedProx based FL model outperforms FedAvg based FL model because of the contribution from the proximal term in the objective function.
As such, the distributed privacy-preserving FedProx based FL platform cam make accurate predictions in the highly dynamic IoV networks 
and can be immensely useful for ADAS and other IoV applications.

\subsubsection{Robustness of the Proposed FL Platform}
\begin{figure*}[!t]
\begin{minipage}{0.33\textwidth}
    \centering
    \includegraphics[width=\textwidth]{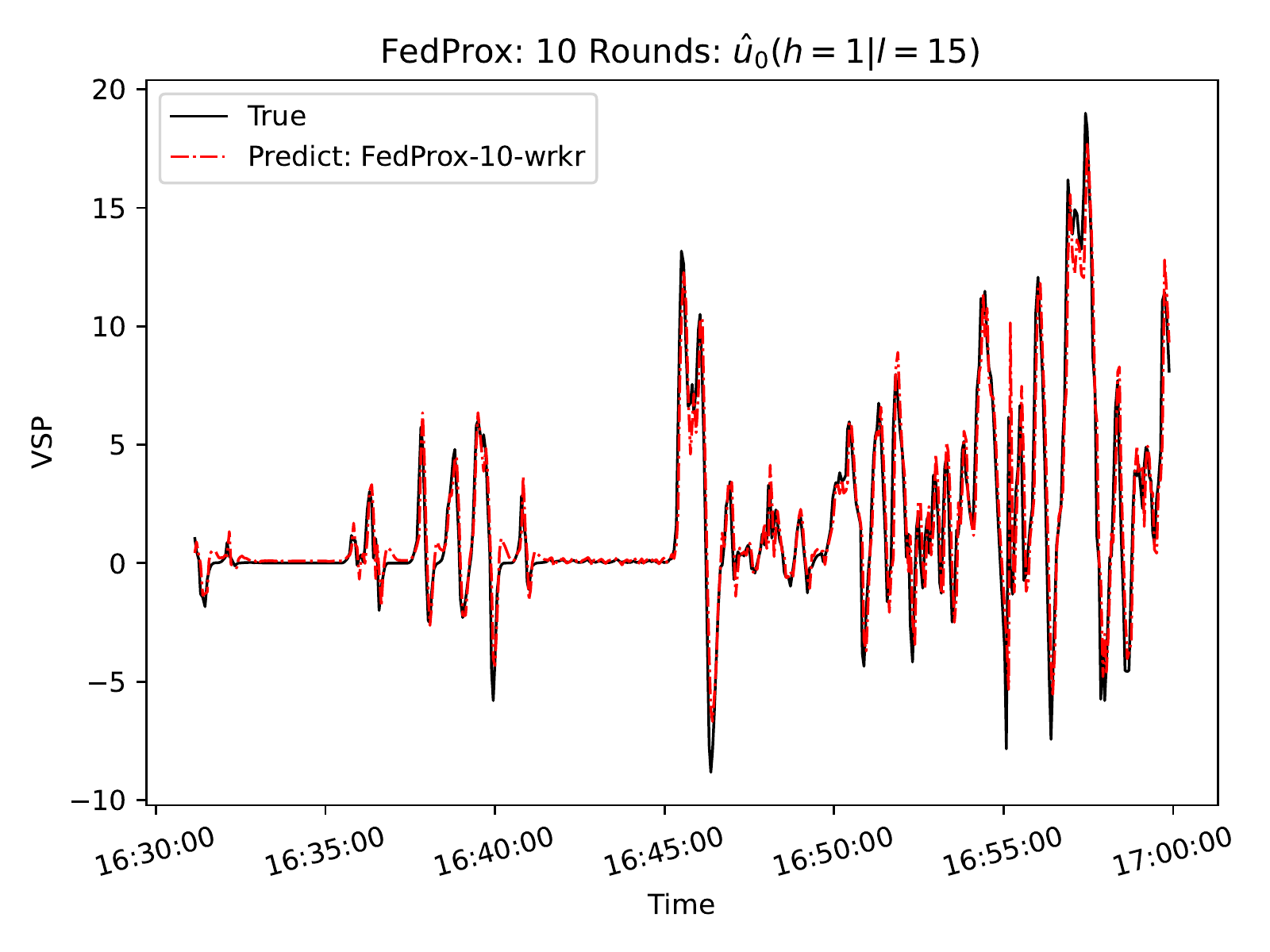} 
    \caption{Predicted VSP for vehicle $1$' with $h=1$}
    \label{predVSP_agent_1_h_1}
\end{minipage} \hspace{0.01in}
\begin{minipage}{0.33\textwidth}
    \centering
    \includegraphics[width=\textwidth]{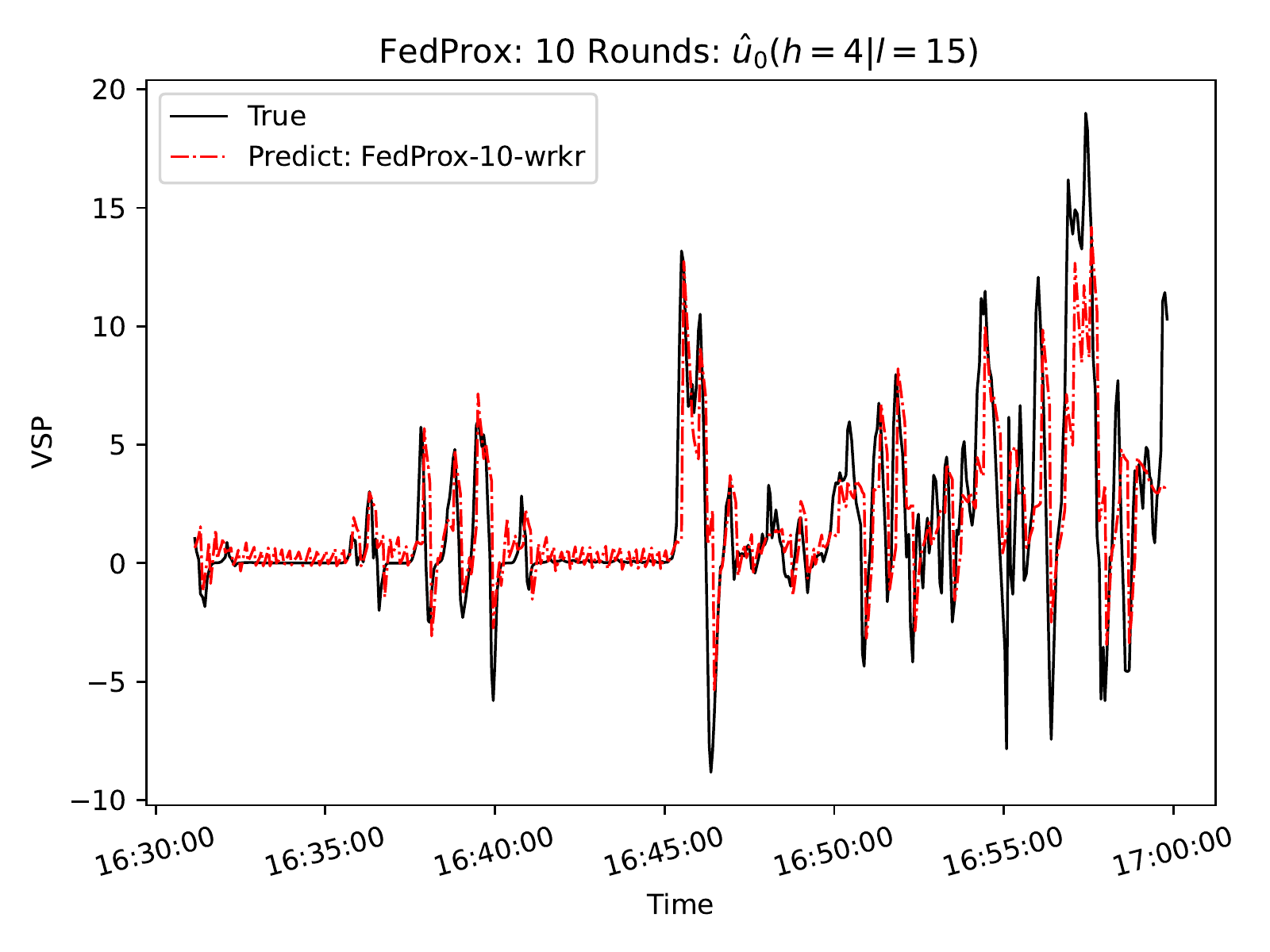} 
    \caption{Predicted VSP for vehicle $1$ with $h=4$}
    \label{predVSP_agent_1_h_4}
\end{minipage} \hspace{0.01in}
\begin{minipage}{0.33\textwidth}
    \centering
    \includegraphics[width=\textwidth]{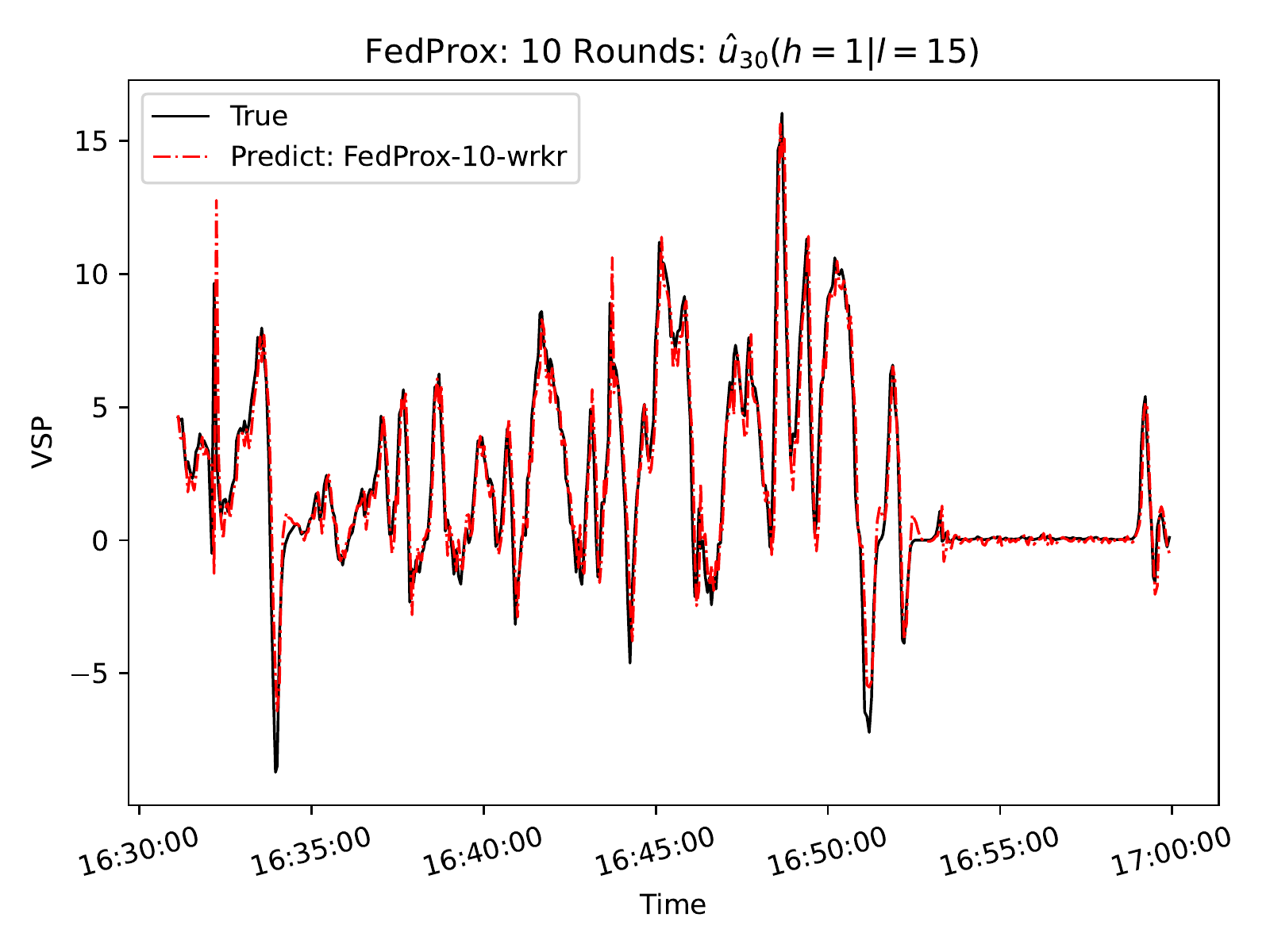} 
    \caption{Predicted VSP for vehicle $30$ with $h=1$}
    \label{predVSP_agent_30_h_1}
\end{minipage}
\begin{minipage}{0.33\textwidth}
    \centering
    \includegraphics[width=\textwidth]{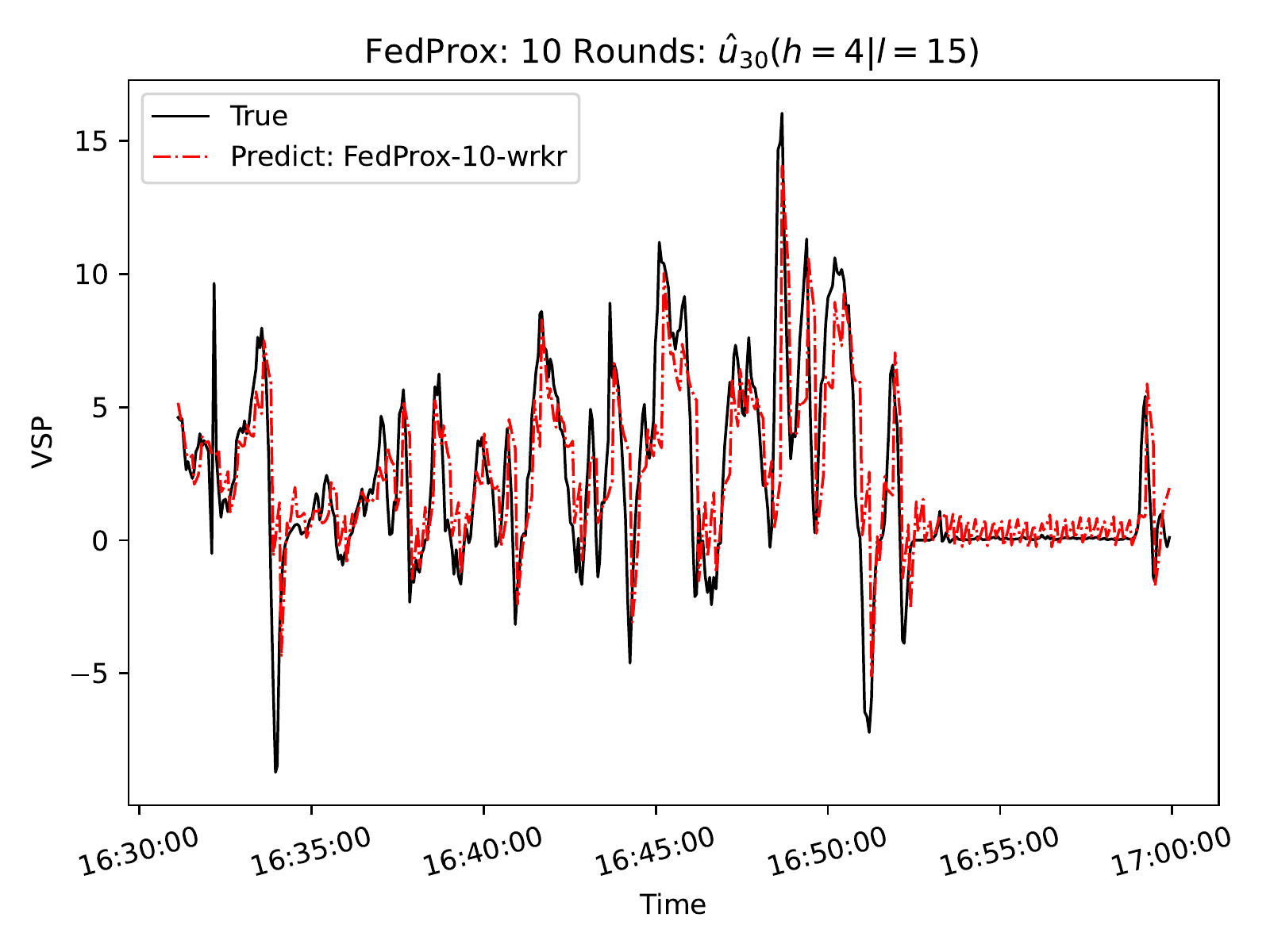} 
    \caption{Predicted VSP for vehicle $30$ with $h=4$}
    \label{predVSP_agent_30_h_4}
\end{minipage} \hspace{0.01in}
\begin{minipage}{0.33\textwidth}
    \centering
    \includegraphics[width=\textwidth]{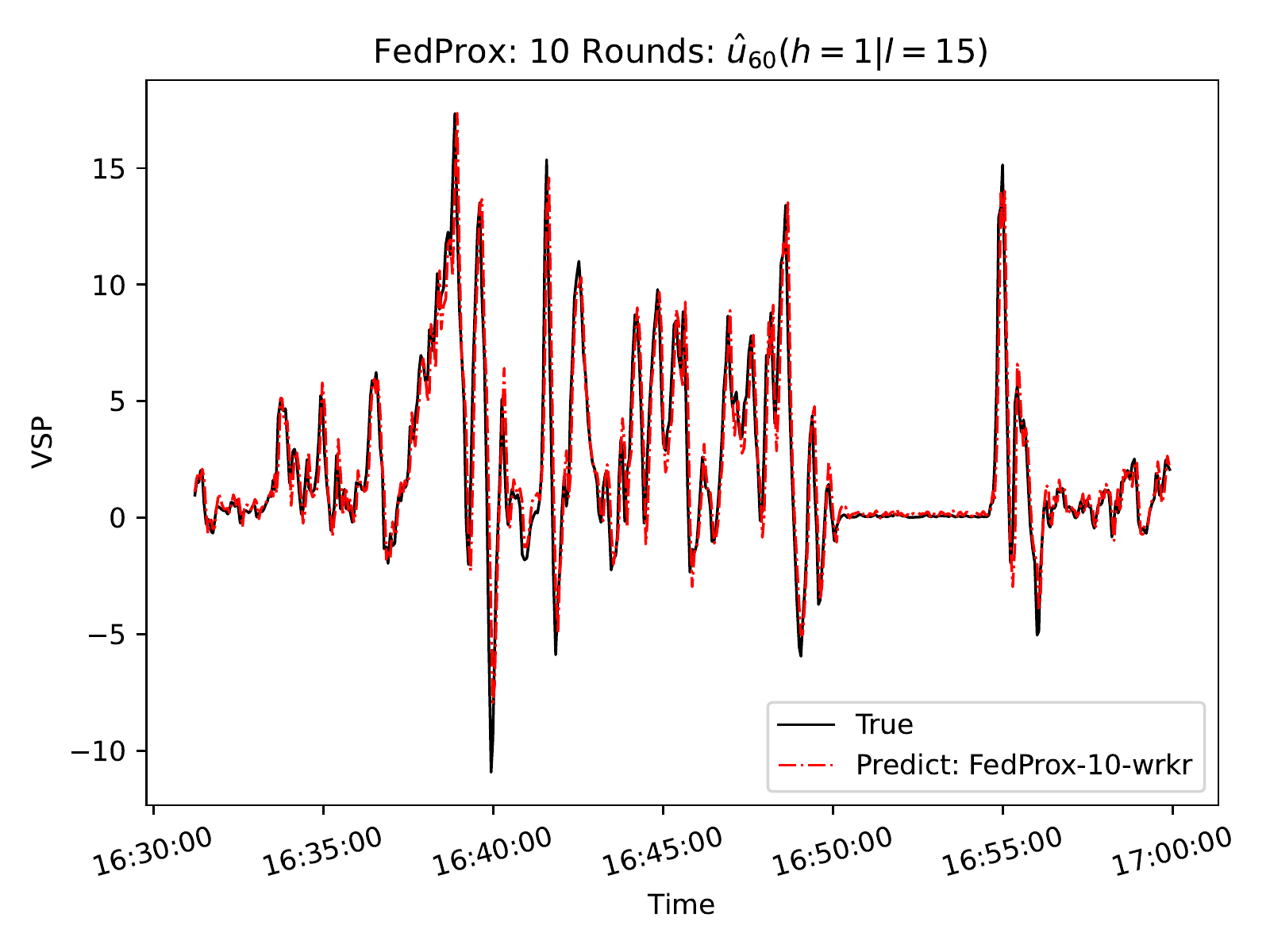} 
    \caption{Predicted VSP for vehicle $60$' with $h=1$}
    \label{predVSP_agent_60_h_1}
\end{minipage} \hspace{0.01in}
\begin{minipage}{0.33\textwidth}
    \centering
    \includegraphics[width=\textwidth]{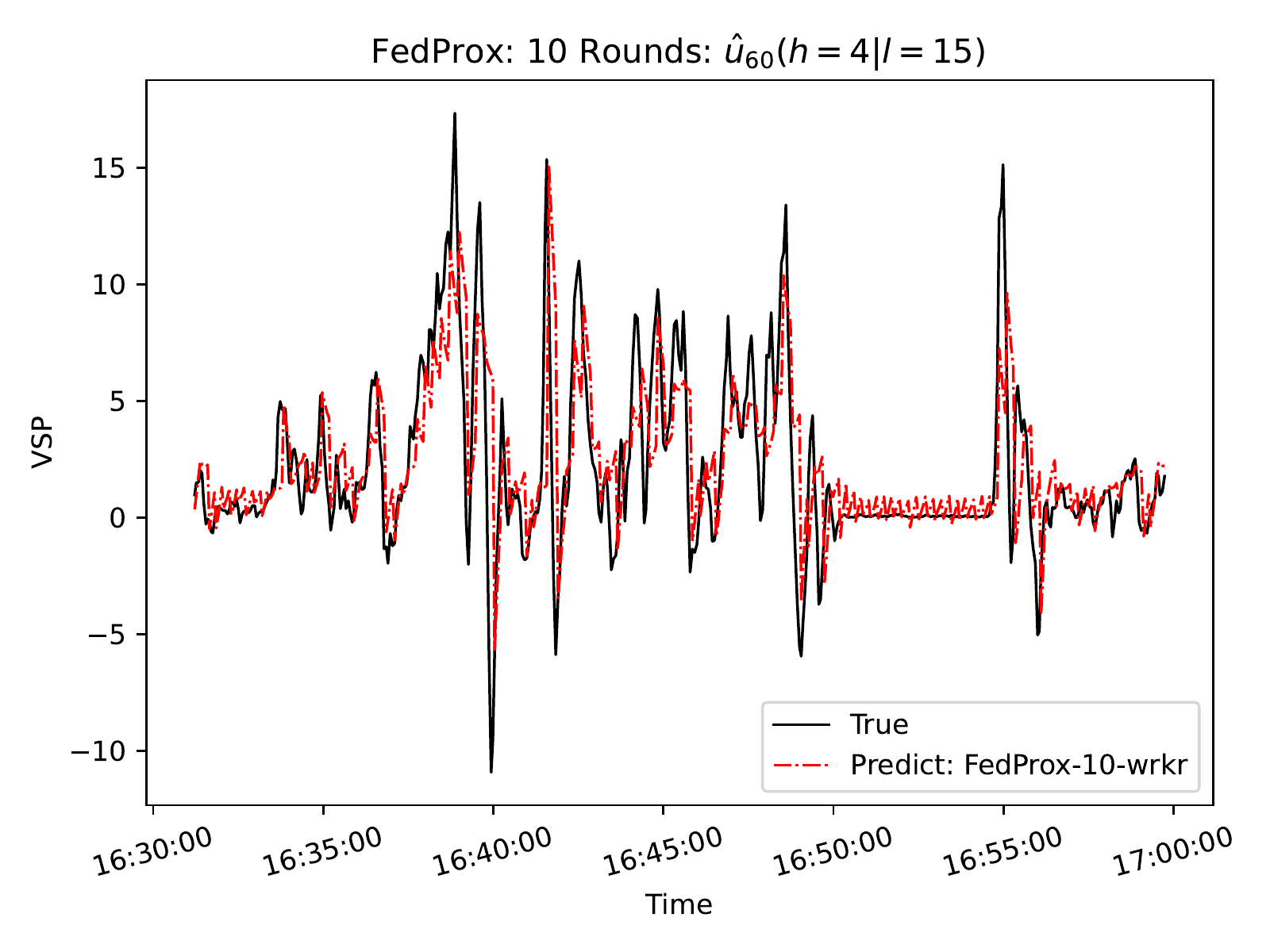} 
    \caption{Predicted VSP for vehicle $60$' with $h=4$}
    \label{predVSP_agent_60_h_4}
\end{minipage}
\end{figure*}

The trained global model can be used by any vehicle to perform their tasks. We show how accurate the model performs.
To that end, we tested our trained global model using the unseen test data samples of all $61$ vehicles. 
Figs. \ref{predVSP_agent_1_h_1} - \ref{predVSP_agent_60_h_4} validate that the proposed distributed FL solution works well for all vehicles.
In addition, our FL solutions ensure that the global model can be trained in an online fashion.
Therefore, our distributed FL solutions are feasible for real-time IoV applications. Thousands of the on-road vehicles can benefit from our FL solutions while only a subset of vehicles need to perform model training.

\subsubsection{Impact of Forecast Window Size}

\iftrue
\begin{figure}[!t]
    \centering
    \includegraphics[width=0.45\textwidth]{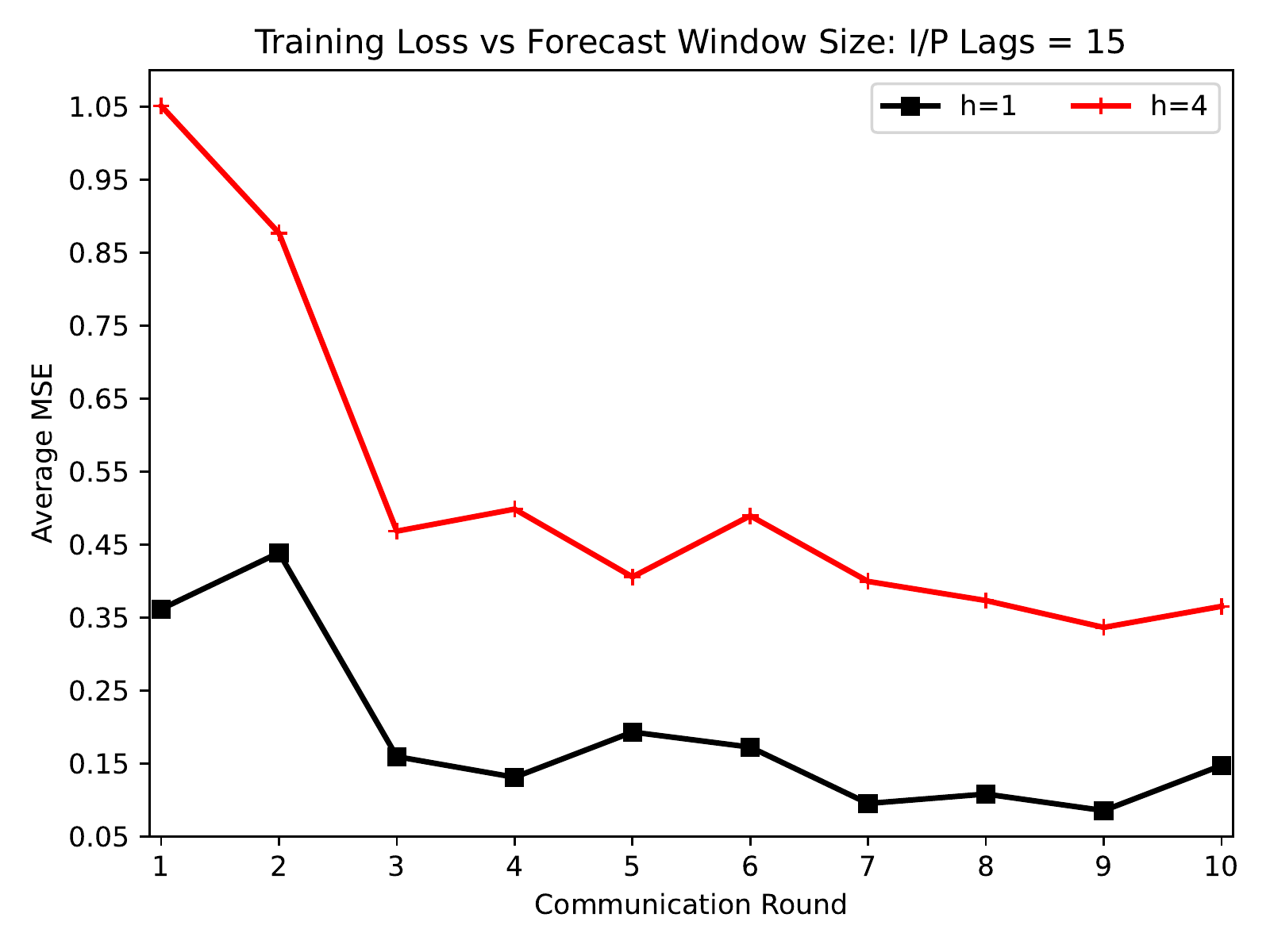}
    \caption{Training loss variation with respect to forecast window size}
    \label{TrainingLoss}
\end{figure}
\fi

Finally, we illustrate the impact of the forecast window size $h$ with a fixed lag of $l=15$. 
As the $h$ increases, the agents need more time to predict the future VSP.
This essentially means that the probability of making an erroneous prediction increases.
We first take $h=1$ and $h=4$ as examples to show the behavior of the average training loss with respect to the forecast window size. 
As shown in Fig. \ref{TrainingLoss},
the average training loss for $h=1$ is less than that for $h=4$ over all communication rounds. 
Furthermore, as the global communication round increases, the model is exposed to more training samples of the diverse agents.
As such, the average training loss exhibits the decreasing tendency.

\begin{table}[!t]
\centering
\small
\caption{\textbf{Average MSE in All $61$ agents Test Data}}
\begin{tabular}{|K{1.4cm}|K{2.2cm}|K{2.6cm}|}
\hline
    \textbf{Forecast Window} $h$ & \textbf{Proposed FL Solution} & \textbf{Proposed Platform + FedAvg} \\ \hline
    $1$ & 2.5784784 & 2.591314118 \\ \hline
    $2$ &  4.63190712 & 4.670199372 \\ \hline
    $3$ & 6.85608755 & 7.09665202 \\ \hline
    $4$ & 9.10875644 & 9.55388634 \\ \hline
    $5$ & 11.72196015 & 11.89405537 \\ \hline
    $6$ & 13.57525046 & 13.39627498 \\ \hline
    $7$ & 14.17268583 & 13.95693221 \\ \hline 
    $8$ & 15.43924312 & 15.44841811 \\ \hline 
    $9$ & 16.07935477 & 16.10373348 \\ \hline 
    $10$ & 16.26966842 & 16.83335492 \\ \hline 
    $11$ & 16.85294033 & 16.56411971 \\ \hline 
    $12$ & 16.27049288 & 16.5502178 \\ \hline 
\end{tabular}
\label{Avg_Test_MSE_All_agents}
\end{table}

We next show average training loss for 
all forecast window sizes using our proposed vehicular FL solutions amalgamating with FedProx and FedAvg. Table \ref{Avg_Test_MSE_All_agents} shows that the prediction accuracy degrades as the forecast window size $h$ increases. 
Therefore, one should carefully choose the forecast window size based on the required accuracy by the applications such as velocity and VSP prediction.
Furthermore, FedProx works better than FedAvg. Accordingly, FedProx based FL solution is more suitable for IoV networks.

\section{Conclusion}
\label{Sec_Conclusion}
This paper proposes a mobility, communication and computation aware integrated FL platform for vehicular environment.  
Leveraging V2X communication, we have devised the I2V/V2I communication platform to exploit on-road vehicles as learning agents.
To maximize local model training time, the proposed FL platform optimizes downlink and uplink radio resource allocations to minimize model transmission delay. 
Moreover, as vehicles have different computation power and dataset, our FL solution tolerates partial works of the agents, i.e., vehicle agents perform heterogeneous local model training, which makes it more suitable for delay-constrained applications in mobile networks.
Using real-world and causal training datasets, we have shown that the proposed FL algorithms can be implemented in an online fashion.
Our simulation results show that the proposed FL platform is robust and can deliver near ground truth velocity and VSP predictions.

\bibliography{Reference}

\begin{thebibliography}{10}
\providecommand{\url}[1]{#1}
\csname url@samestyle\endcsname
\providecommand{\newblock}{\relax}
\providecommand{\bibinfo}[2]{#2}
\providecommand{\BIBentrySTDinterwordspacing}{\spaceskip=0pt\relax}
\providecommand{\BIBentryALTinterwordstretchfactor}{4}
\providecommand{\BIBentryALTinterwordspacing}{\spaceskip=\fontdimen2\font plus
\BIBentryALTinterwordstretchfactor\fontdimen3\font minus
  \fontdimen4\font\relax}
\providecommand{\BIBforeignlanguage}[2]{{%
\expandafter\ifx\csname l@#1\endcsname\relax
\typeout{** WARNING: IEEEtran.bst: No hyphenation pattern has been}%
\typeout{** loaded for the language `#1'. Using the pattern for}%
\typeout{** the default language instead.}%
\else
\language=\csname l@#1\endcsname
\fi
#2}}
\providecommand{\BIBdecl}{\relax}
\BIBdecl

\bibitem{mcmahan2017communication}
B.~McMahan \emph{et~al.}, ``Communication-efficient learning of deep networks
  from decentralized data,'' in \emph{Artificial intelligence and statistics},
  2017.

\bibitem{9311906}
S.~Hosseinalipour \emph{et~al.}, ``From federated to fog learning: Distributed
  machine learning over heterogeneous wireless networks,'' \emph{IEEE Commun.
  Mag.}, vol.~58, no.~12, pp. 41--47, Dec. 2020.

\bibitem{zeng2021}
T.~Zeng \emph{et~al.}, ``Multi-task federated learning for traffic prediction
  and its application to route planning,'' in \emph{Proc. of IEEE IV}, July
  2021.

\bibitem{kong2021fedvcp}
X.~Kong, H.~Gao, G.~Shen, G.~Duan, and S.~K. Das, ``Fedvcp: A
  federated-learning-based cooperative positioning scheme for social internet
  of vehicles,'' \emph{IEEE Trans. Computational Social Syst.}, vol.~9, no.~1,
  pp. 197--206, 2022.

\bibitem{liu2020client}
L.~Liu, J.~Zhang, S.~Song, and K.~B. Letaief, ``Client-edge-cloud hierarchical
  federated learning,'' in \emph{Proc. IEEE ICC}, June 2020.

\bibitem{zhou2021two}
X.~Zhou, W.~Liang, J.~She, Z.~Yan, and K.~I.-K. Wang, ``Two-layer federated
  learning with heterogeneous model aggregation for 6g supported internet of
  vehicles,'' \emph{IEEE Trans. Vehicular Technol.}, vol.~70, no.~6, pp.
  5308--5317, 2021.

\bibitem{MLSYS2020_38af8613}
T.~Li \emph{et~al.}, ``Federated optimization in heterogeneous networks,'' in
  \emph{Proc. of Machine Learn. and Syst.}, vol.~2, 2020, pp. 429--450.

\bibitem{3GPP_TR_38_901}
``{\textit{3rd Generation Partnership Project; Technical Specification Group
  Radio Access Network; Study on channel model for frequencies from 0.5 to 100
  GHz}},'' 3GPP TR 38.901 V16.1.0, Release 16, Dec. 2019.

\bibitem{3GPP_TR_38_886}
``{\textit{3rd Generation Partnership Project; Technical Specification Group
  Radio Access Network; V2X Services based on NR; User Equipment (UE) radio
  transmission and reception}},'' 3GPP TR 38.886 V0.5.0, Release 16, Feb. 2020.

\bibitem{sesia2011lte}
S.~Sesia, I.~Toufik, and M.~Baker, \emph{LTE-the UMTS long term evolution: from
  theory to practice}.\hskip 1em plus 0.5em minus 0.4em\relax John Wiley \&
  Sons, 2011.

\bibitem{grant2009cvx}
M.~Grant, S.~Boyd, and Y.~Ye, ``cvx users’ guide,'' \emph{online: http://www.
  stanford. edu/\~{} boyd/software. html}, 2009.

\bibitem{kuhn1955hungarian}
H.~W. Kuhn, ``The hungarian method for the assignment problem,'' \emph{Naval
  research logistics quarterly}, vol.~2, no. 1-2, pp. 83--97, 1955.

\bibitem{herrera2010evaluation}
J.~C. Herrera \emph{et~al.}, ``Evaluation of traffic data obtained via
  {GPS}-enabled mobile phones: The mobile century field experiment,''
  \emph{Transport. Research Part C: Emerg. Technol.}, vol.~18, no.~4, pp.
  568--583, 2010.

\bibitem{koupal2005moves2004}
J.~Koupal \emph{et~al.}, ``Moves 2004: Energy and emission inputs draft
  report,'' \emph{US Environmental Protection Agency, Report No.
  EPA420-P-05-003}, 2005.

\end{thebibliography}

\end{document}